\documentclass[lettersize,journal]{IEEEtran}
\usepackage{amsmath,amsfonts}
\usepackage{algorithmic}
\usepackage{algorithm}
\usepackage{array}
\usepackage[caption=false,font=normalsize,labelfont=sf,textfont=sf]{subfig}
\usepackage{textcomp}
\usepackage{stfloats}
\usepackage{url}
\usepackage{verbatim}
\usepackage{graphicx}

\usepackage{cite}
\usepackage{color}
\usepackage{multirow}
\usepackage{makecell}
\usepackage{arydshln}

\usepackage[capitalize]{cleveref}
\crefname{section}{Sec.}{Secs.}
\crefname{table}{Tab.}{Tabs.}
\Crefname{figure}{Fig.}{Figs.}
\Crefname{equation}{Eq.}{Eqs.}
\Crefname{algorithm}{Alg.}{Algs.}

\begin{document}

\title{Don't Stop Learning: Towards Continual Learning for the CLIP Model}

\author{Yuxuan~Ding,~\IEEEmembership{}
	Lingqiao~Liu,~\IEEEmembership{}
	Chunna~Tian,~\IEEEmembership{}
	Jingyuan~Yang~\IEEEmembership{}
	and~Haoxuan~Ding~\IEEEmembership{}
	\thanks{
		(\textit{Corresponding author: Chunna Tian, Lingqiao Liu.})
		
		Yuxuan Ding and Chunna Tian are with the School of Electronic Engineering, Xidian University, Xi'an 710071, China (e-mail: yxding@stu.xidian.edu.cn; chnatian@xidian.edu.cn).
		
		Lingqiao Liu is with the Australian Institute for Machine Learning, the University of Adelaide, Adelaide 5005, Australia (e-mail: lingqiao.liu@adelaide.edu.au).
		
		Jingyuan Yang is with the College of Computer Science and Software Engineering, Shenzhen University, China (e-mail: jyyang@szu.edu.cn).
		
		Haoxuan Ding is with the Unmanned System Research Institute, Northwestern Polytechnical University, Xi'an 710072, China (e-mial: haoxuan.ding@mail.nwpu.edu.cn) 
		
	}
}


\markboth{}
{Shell \MakeLowercase{\textit{et al.}}: A Sample Article Using IEEEtran.cls for IEEE Journals}


\maketitle

\begin{abstract}
	The Contrastive Language-Image Pre-training (CLIP) Model is a recently proposed large-scale pre-train model which attracts increasing attention in the computer vision community. Benefiting from its gigantic image-text training set, the CLIP model has learned outstanding capabilities in zero-shot learning and image-text matching. To boost the recognition performance of CLIP on some target visual concepts, it is often desirable to further update the CLIP model by fine-tuning some classes-of-interest on extra training data. This operation, however, raises an important concern: will the update hurt the zero-shot learning or image-text matching capability of the CLIP, i.e., the catastrophic forgetting issue? If yes, could existing continual learning algorithms be adapted to alleviate the risk of catastrophic forgetting? To answer these questions, this work conducts a systemic study on the continual learning issue of the CLIP model. We construct evaluation protocols to measure the impact of fine-tuning updates and explore different ways to upgrade existing continual learning methods to mitigate the forgetting issue of the CLIP model. Our study reveals the particular challenges of CLIP continual learning problem and lays a foundation for further researches. Moreover, we propose a new algorithm, dubbed Learning without Forgetting via Replayed Vocabulary (VR-LwF), which shows exact effectiveness for alleviating the forgetting issue of the CLIP model. 
\end{abstract}


\section{Introduction}
\begin{figure*}[t]
	\centering
	\begin{minipage}[t]{0.48\textwidth}
		\centering
		\includegraphics[width=\textwidth]{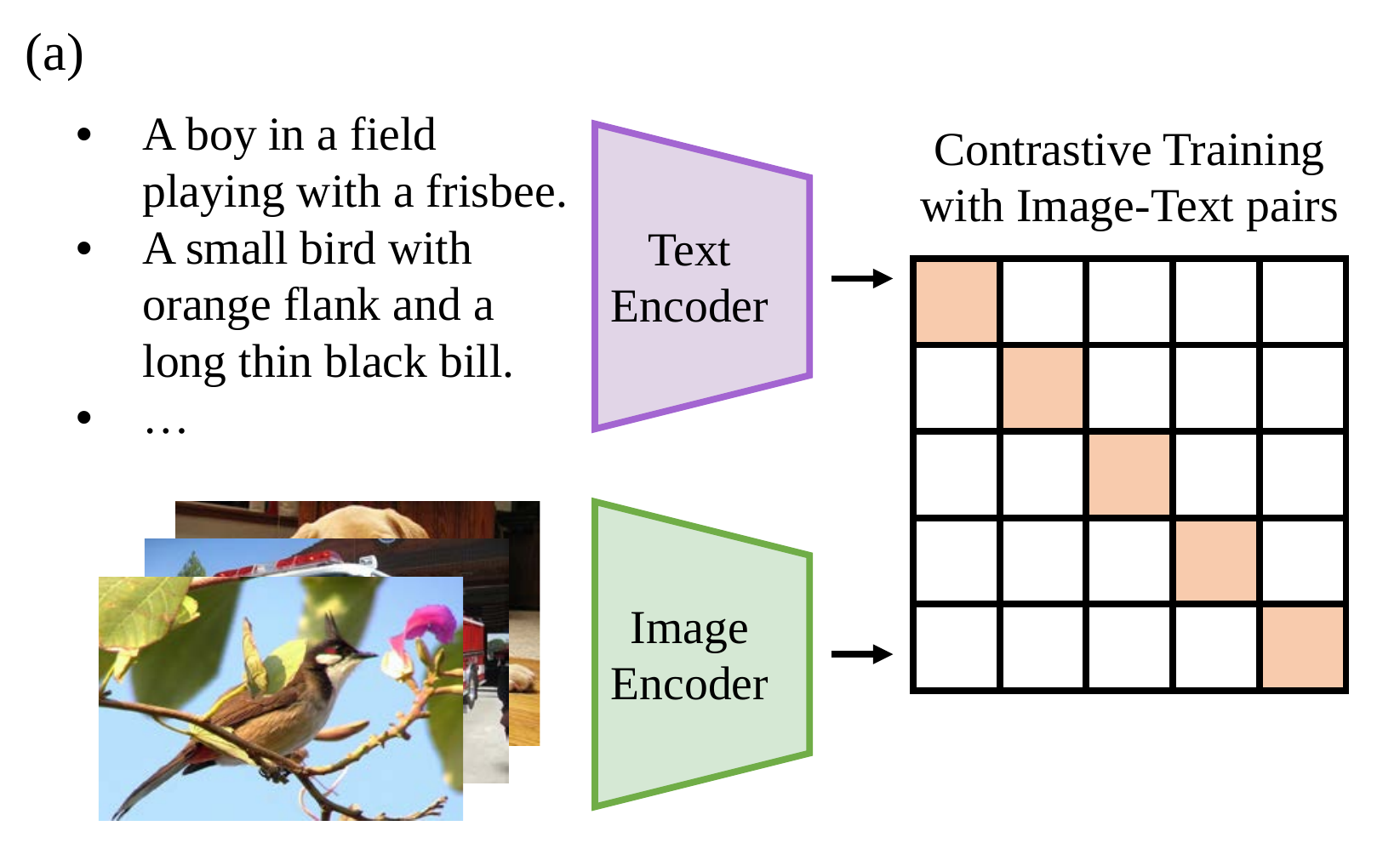}
	\end{minipage}
	\begin{minipage}[t]{0.5\textwidth}
		\centering
		\includegraphics[width=\textwidth]{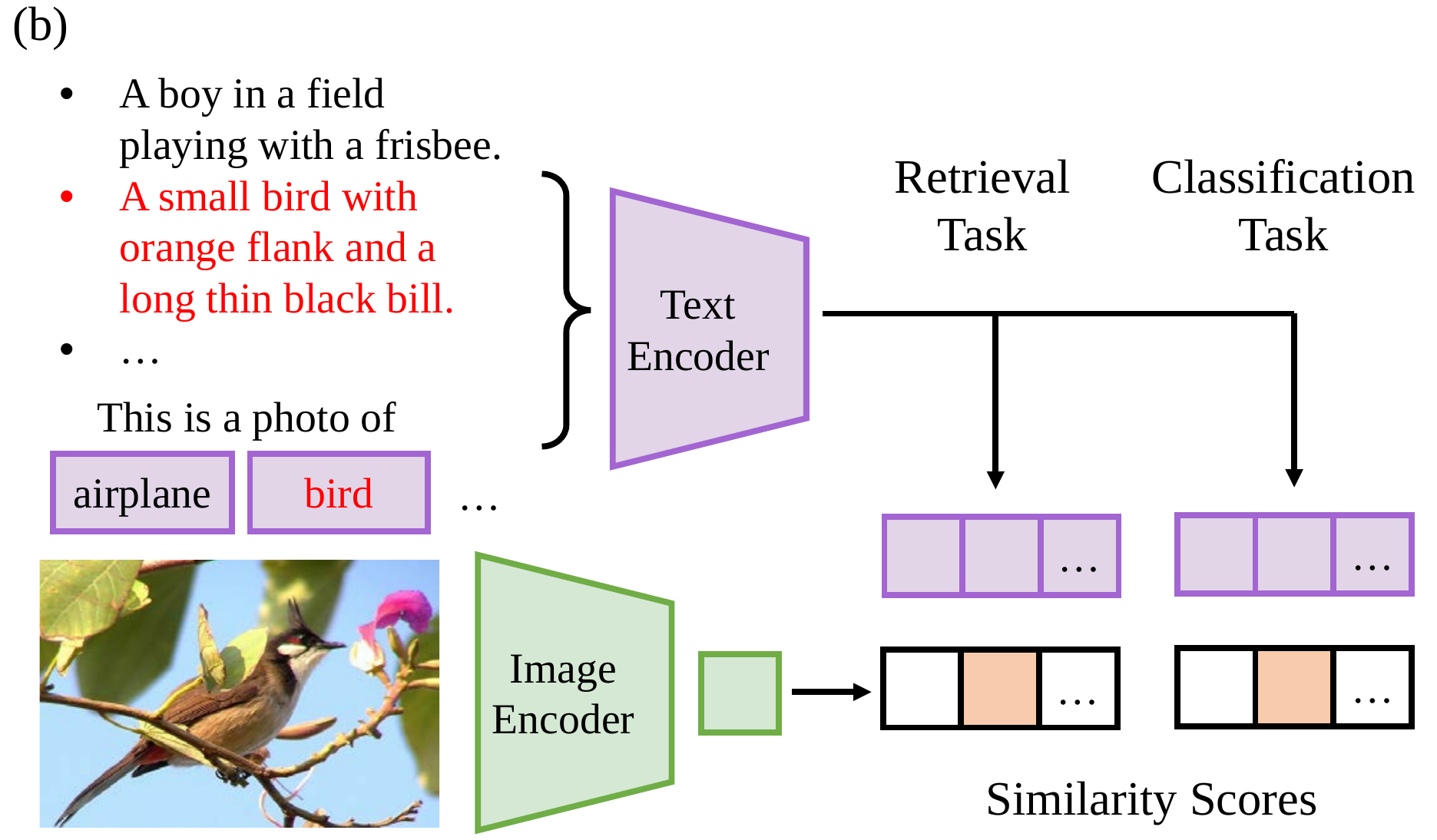}
	\end{minipage}
	\caption{The workflow of the CLIP model. (a) The training of CLIP is conducted by a huge dataset of noisy image-text pairs. (b) At zero-shot inference, the input of the text encoder is retrieval captions or text prompts. Image and text embeddings are generated by the encoders and compute the similarities. The most similar match can be seen as the prediction.}
	\label{CLIP_Frame}
\end{figure*}

Motivated by the success of pre-trained language models like Bidirectional Encoder Representation from Transformers (BERT) \cite{BERT} and Generate Pre-Training Model (GPT) \cite{GPT} in natural language processing, the vision-language community has started to embrace the idea of pre-training large neural networks on huge datasets. Various methods, such as ViLBERT \cite{VilBERT} and LXMERT \cite{Lxmert}, have been developed in recent years. Contrastive Language-Image Pre-training (CLIP) \cite{CLIP} is one of the latest advances in this direction. CLIP is optimized by the contrastive loss over a training set of 400 million noisy image-text pairs crawled from the Internet. The final model demonstrates unprecedented zero-shot classification capacity. The training and inference pipelines are illustrated in \Cref{CLIP_Frame}. The zero-shot inference is applied by text prompt technology, the name of a class will be transformed into a sentence, such as ``this is a photo of [Class Name].", and then contrast with images (details are explained in \Cref{CLIP}). Under the circumstances, the CLIP tackles any classes which can be described by language and outperforms state-of-the-art fully-supervised baselines on 16 datasets without using any labeled data. The model quickly sparks attention since its release, which has been successfully applied in zero-shot detection \cite{DetCLIP}, open set detection \cite{OpenSetCLIP} and video retrieval \cite{VRCLIP, CLIP4Clip, CLIP2Video}.

However, even with its gigantic volume of training data and wide coverage of visual concepts, the CLIP model still cannot attain satisfying performance in some visual categories. A practical solution is to fine-tune CLIP on extra training examples collected for categories where CLIP falls short of expectation. However, this na\"ive solution raises a concern on \textbf{whether fine-tuning CLIP would hurt its zero-shot learning and/or image-text matching capabilities}. This concern is related to the catastrophic forgetting problem that has been intensively studied in continual learning with the context of image classification. Thus a natural follow-up question is \textbf{whether existing continual learning approaches could help alleviate the potential forgetting issue}.


Those two questions motivate this paper. We conduct a systematic study on the continual learning problem of the CLIP model, focusing on the impact of additional fine-tuning on the zero-shot and text-image matching performance of the CLIP model.\footnote{We use the "ViT-B/32" CLIP model in this paper.} Our study shows that the CLIP model also suffers from the catastrophic forgetting problem: the zero-shot learning and image-text matching capabilities would be significantly damaged after fine-tuning. We further investigate whether existing continual learning approaches can be adapted as a solution. We explore this possibility by modifying four representative continual learning approaches, \emph{i.e.}, LwF \cite{LwF}, GeoDL \cite{GeoDL}, IMM \cite{IMM}, and RKR \cite{RKR}, from three commonly used families of continual learning. Moreover, we propose a new solution to address the forgetting issue of the CLIP model. Our key idea is to enforce the prediction logits for a set of randomly synthesized classes before and after the model update. Despite its simplicity, this approach demonstrates excellent performance in our experimental study and outperforms the direct extensions of existing continual learning approaches. It can mitigate the forgetting issue while making the CLIP still be benefited from fine-tuning.

To sum up, the main contributions of this paper are threefold:

\begin{itemize}
	\item We present a systematic study of continual learning problem for the CLIP model. An evaluation protocol is designed to assess the learning and forgetting after fine-tuning the original CLIP.
	\item We explore the extension of popular existing continual learning algorithms on the CLIP model.
	\item We propose a simple-but-effective solution named Learning without Forgetting via Replayed Vocabulary (VR-LwF), which largely alleviates the catastrophic forgetting after CLIP fine-tuning. 
\end{itemize}

\section{Background and Related Work}
\subsection{The CLIP Model}
\label{CLIP}
Motivated by the success of pre-trained language models in NLP, \emph{e.g.}, BERT \cite{BERT}, several vision-language-based pre-training models have been proposed recently. Early attempts \cite{videobert, UNITER, VisualBERT, VilBERT, 12in1, Lxmert} usually learn with a strategy that is similar to the mask language modeling in BERT. More recently, CLIP \cite{CLIP}, and several similar works \cite{ALIGN, WenLan} adopt the contrastive learning strategy and a dual-tower structure to build the model. They utilize an image encoder and a text encoder to process the input images and the input sentence separately. After obtaining their embeddings, a contrastive-learning-alike loss is utilized to encourage correct matching across modalities.

Although being trained for image-text matching, the CLIP model shows an outstanding capability in zero-shot classification. To perform zero-shot classification, CLIP employs a prompt to convert the classification problem into text matching. For example, the candidate class names are filled into a textual prompt like ``this is a photo of [Class Name].", then the prompt text is encoded to match the visual feature. By comparing similarities between image and text embeddings, the probability of assigning an image to a candidate class is calculated as
\begin{align}
	\label{clip_prob}
	&p\left( {y = c|{\bf{x}}} \right) = \frac{{\exp \left( {\tau  \cdot \left\langle {{\bf{f}},{{\bf{t}}^c}} \right\rangle }  \right)}}{{\sum\nolimits_{j = 1}^K {\exp \left( {\tau  \cdot \left\langle {{\bf{f}},{{\bf{t}}^j}} \right\rangle } \right)} }},  \\ 
	&{\bf{f}} = {{\mathop{\rm CLIP}\nolimits} _i}\left( {{\bf{x}};{{\bf{\theta }}_i}} \right), \\ 
	&{{\bf{t}}^c} = {{\mathop{\rm CLIP}\nolimits} _t}\left( {{\mathop{\rm Prompt}\nolimits} \left( {{y^c}} \right);{{\bf{\theta }}_t}} \right), \label{text_encode}
\end{align}
where $\mathbf{x}$ and $y^c$ denote the input image and class name of the $c$-th class, respectively. ${{\mathop{\rm CLIP}\nolimits} _i}\left(  \cdot  \right)$ and ${{\mathop{\rm CLIP}\nolimits} _t}\left(  \cdot  \right)$ are image encoder and text encoder, ${\bf{f}}$ and ${{\bf{t}}^c}$ are derived visual and textual embeddings. ${\mathop{\rm Prompt}\nolimits} \left( {{y^c}} \right)$ denotes the prompt extension of class name $y^c$. For simplicity, we omit ${\mathop{\rm Prompt}\nolimits} \left( \cdot \right)$ and use ${\bf y}^c$ in the following sections. $\tau$ is a logit scaling factor and $\left\langle { \cdot , \cdot } \right\rangle $ is the inner product between two normalized embedding vectors. The framework of original CLIP is illustrated in \Cref{CLIP_Frame}.

\subsection{Continual Learning}

Continual learning (CL) aims at solving the catastrophic forgetting problem \cite{forgetting}, which refers to the performance drop of preceding tasks when learning on a new task, \emph{i.e.}, forgetting previously acquired knowledge. Continual learning has been considerably interested in many areas, such as classification \cite{LC}, reinforcement learning \cite{IRL}, and online learning \cite{OIC}. Main solutions can be grouped into four categories: adaptive-plasticity-based methods, distillation-based methods, replay-based methods, and architecture-based methods. 

Adaptive-plasticity-based methods learn different levels of plasticity for the model parameters, expecting important parameters of former tasks could be retained as much as possible. EWC \cite{EWC} is a typical study, authors use Fisher information matrix to measure the importance of parameters, the higher Fisher information brings higher penalty on plasticity. But EWC preserves the penalties on all previous tasks, which will cost heavy when task number is large. Online-EWC \cite{OEWC} is a further research, where looses the penalty only on the last old task. MAS \cite{MAS} and SI \cite{SI} estimate the neuron importance by observing the influence of small parameter perturbation respectively on network output and loss value. NPC \cite{NPC} assigns the stable neurons lower learning rates rather than restriction to retain knowledge. IMM \cite{IMM} uses the framework of Bayesian network and proposes two methods, IMM-Mean and IMM-Mode. IMM-Mean simply average the parameters of new and old models, IMM-Mode combines two networks with Laplacian approximation \cite{Laplacian}.

Distillation-based methods usually equip with knowledge distilling technique \cite{KD}. The current model is trained with supervision from both the new task and a teacher model, which is usually the old model. LwF \cite{LwF}, LwM \cite{LwM}, PODNet \cite{Podnet} and MCIL \cite{MCIL} representative works in this family. LwF distills the output logits from the old model. LwM keeps both the Grad-CAM \cite{Grad-cam} attention maps and logits. PODNet distills feature maps with a mixture of different pooling types. PODNet and MCIL both distill knowledge from a exemplar set. But MCIL proposes to learn at exemplar-level, the exemplars are adjusted at each task.

Replay-based methods have to store a small number of samples from the previous tasks \cite{SER, iCARL, Gdumb} or learn to generate synthetic data \cite{Generative,VAECL}. SER \cite{SER} uses reservoir sampling strategy to store the experiences. iCARL \cite{iCARL} builds a fixed $K$ memory budget to store former tasks' exemplars. Samples which is best approximate to the class prototypes are inputted to the budget first. When new classes are added, exemplars at the end of each class will be removed. Approaches using synthetic data also called pseudo-replay methods. DGR \cite{Generative} employs generative adversarial network (GAN) \cite{GAN}. The GAN is updated by current data and replayed synthetic data at each task so as to generate replayed sample in the future task session. CCL-GM \cite{VAECL} is designed with variational autoencoder (VAE) \cite{VAE} architecture, the previous model can be directly used as a generator.

Architecture-based methods modify the network architecture, such as dynamically expanding the model parameters \cite{DER,DEN} or masking different parameters for different tasks \cite{HAT}. It is notable that due to the costing pre-training of CLIP, many continual learning methods are limited to fine-tune the CLIP. Methods with exemplars and regularization methods need the previous training information are not optimal for CLIP continual learning. 

\section{Continual Learning for the CLIP Model}
\subsection{Challenges}
The zero-shot learning capability of CLIP comes from the massive amount of image-text pairs in its training set. However, due to the impossibility of covering all visual concepts in a single training stage, it is not surprising that CLIP fails to produce a satisfactory classification performance on some categories. A practical solution is to fine-tune the CLIP with extra training data for the target classes. However, this may lead to the catastrophic forgetting issue. That is, the zero-shot capability of CLIP may be damaged after fine-tuning. Thus, we may need continual learning methods to overcome this issue.

Compared with the standard-setting in the existing literature of continual learning, the continual learning problem on CLIP (denoted as the CL-CLIP problem hereafter) has the two major differences:
\begin{itemize}
	\item For CL-CLIP problems, we expect to maintain the zero-shot learning capability of the model while making the model still benefit from fine-tuning with extra training data. In contrast, the traditional CL problem tries to maintain the classification capability of previously learned image categories.
	\item The CLIP training is conducted on image-text pairs. There is no explicit ``class'' concept, and the visual patterns covered by CLIP are several orders larger than existing CL settings.  
\end{itemize}
The above differences make the CL-CLIP problem particularly challenging. For example, since in zero-shot classification, the classes are not directly trained at the training stage, the logits difference between a true class and the other classes may be small. Thus a slight change in the model may lead to a wrong prediction, and therefore the model might forget the zero-shot learning capability more easily. The second difference prevents us from directly using replay methods to solve the CL-CLIP problem because we cannot directly store exemplar images for the learned ``classes''. Besides, some adaptive-plasticity-based methods are impracticable, such as EWC\cite{EWC}, SI\cite{SI}, and MAS\cite{MAS}. These methods need previous training procedure to provide information, like Fisher information in EWC and importance estimation in SI and MAS, for continual learning. However, the pre-training of CLIP is costly (400 million data).

\subsection{Evaluation of the CL-CLIP}
\label{protocol}
\begin{figure}[t]
	\centering
	\includegraphics[width=0.48\textwidth]{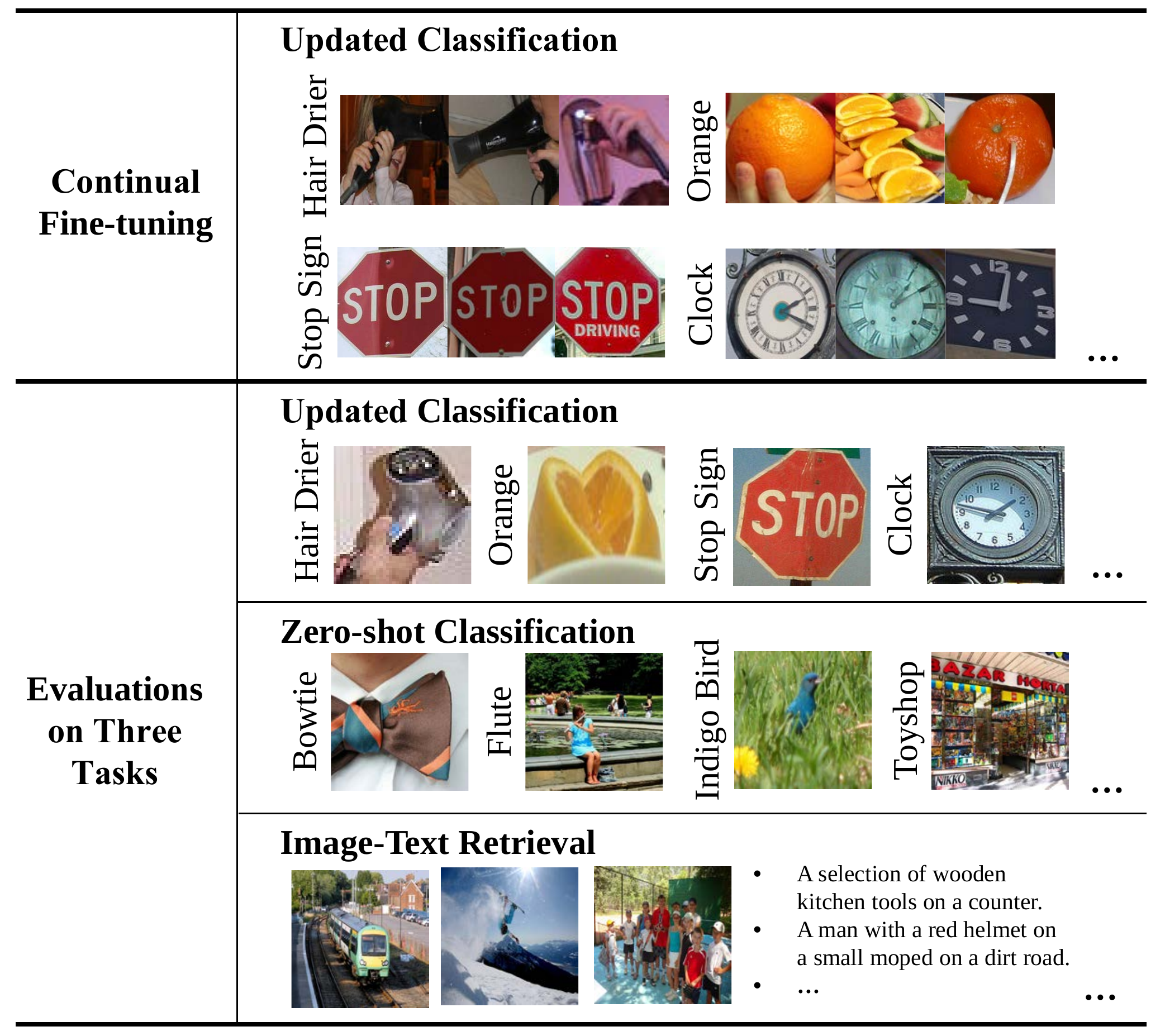}
	\caption{Overview of our proposed evaluation protocols. The CLIP model is fine-tuned with a constructed object classification dataset. After fine-tuning, the updated task, zero-shot classification and image-text retrieval task are evaluated.}
	\label{demo}
\end{figure}

Since the CL-CLIP is a new problem and there is no existing evaluation benchmark\footnote{A recent work \cite{WiSE-FT} also addresses the fine-tuning problem of CLIP, but focuses on improving the out-of-distribution generalization capability of CLIP in the context of zero-shot learning rather than the continual learning problem.}. We conduct a systematic study on the CL-CLIP problem. The goal of a CL-CLIP algorithm is twofold: (1) When fine-tuning on new training data of to-be-enhanced classes, called updated task or updated classes, the classification performance is supposed to be improved. (2) Zero-shot learning performance of the fine-tuned model should be largely maintained or even improved compared to the original CLIP. Hence, in our study, we use additional training data to first update the CLIP model. Then we evaluate the updated model by its performance on updated classes, zero-shot learning, and image-text matching tasks. In other words, there are one training set and three test datasets (demonstrated in \Cref{demo}). In the following part, we elaborate on our evaluation protocol.

\paragraph{Data for Updated Task}
We use MSCOCO \cite{COCO} to construct the training data of the updated classes. 80-way objects are cropped by their bounding-boxes to build the classification task. To be specific, training images are from the standard MSCOCO training set, while the test set is from the images of the matching task described below. There are 596,974 cropped training images and 35,360 testing images in total. We also preserve a validation set from the 5000 validation images in Karpathy's split \cite{karpathy} of MSCOCO.

\paragraph{Data for Zero-shot Learning}
We use tiered-ImageNet \cite{teired-ImageNet} dataset, a subset of ImageNet \cite{Imagenet}, as the evaluation set for zero-shot learning performance. 351 classes with 100 samples in each class are utilized for zero-shot evaluation. Since our target is to examine the CL-CLIP methods, not pursuing the careful prompt engineering, we use a basic prompt text: ``this is a photo of [Class Name]." The rest 160 classes of tiered-ImageNet are reserved for validation.

\paragraph{Data for Image-text Matching}
Besides classification, MSCOCO \cite{COCO} is also used as an image-text retrieval dataset. Three are 5,000 testing images and 25,000 captions (Karpathy's test split), which are usually tested by averaging over five folds of 1,000 images. Note that the updated classes are included in the retrieval texts. We deliberately maintain this overlap and expect a perfect CL-CLIP algorithm can lead to improved performance on image-text matching since the modeling of the overlapped visual concepts is improved via fine-tuning.

\paragraph{One-Session and Multi-Session Training} 
Apart from the dataset, two fine-tuning settings are designed for evaluating CL-CLIP methods, \textbf{One-Session Training (OST)} and \textbf{Multi-Session Training (MST)}. One-session training only updates the model with one more task, where all training data is accessible at the same time. Multi-session training provides training classes sequentially, which is similar to the typical setting of continual learning. Specifically, 80 classes are divided into 8 sessions (the splits are listed in \Cref{MST_split}). Data from each session are used to sequentially update the model. This is regarded as a complex situation of CL-CLIP.

\begin{table}[t]
	\centering
	\small
	\caption{The session splits for MSCOCO multi-session training. The prompted names are located in the second column. The amounts of training and testing samples are listed in the third column.}
	\label{MST_split}
	\resizebox{0.45\textwidth}{!}{%
		\begin{tabular}{c|m{50mm}|c}
			\hline
			Session & \makecell[c]{Class Names} & Number of Samples \\ \hline
			1 & person, bicycle, car, motorcycle, airplane, bus, train, truck, boat, traffic light. & 258,719 / 14,886 \\ \hline
			2 & fire hydrant, stop sign, parking meter, bench, bird, cat, dog, horse, sheep, cow. & 41,024 / 2,678 \\ \hline
			3 & elephant, bear, zebra, giraffe, backpack, umbrella, handbag, tie, suitcase, frisbee. & 45,049 / 2,643 \\ \hline
			4 & skis, snowboard, sports ball, kite, baseball bat, baseball glove, skateboard, surfboard, tennis racket,   bottle. & 49,939 / 3,048 \\ \hline
			5 & wine glass, cup, fork, knife, spoon, bowl, banana, apple, sandwich, orange. & 62,360 / 3,801 \\ \hline
			6 & broccoli, carrot, hot dog, pizza, donut, cake, chair, couch, potted plant, bed. & 65,932 / 3,947 \\ \hline
			7 & dining table, toilet, tv, laptop, mouse, remote, keyboard, cell phone, microwave, oven. & 36,776 / 2,127 \\ \hline
			8 & toaster, sink, refrigerator, book, clock, vase, scissors, teddy bear, hair drier, toothbrush. & 37,175 / 2,230 \\ \hline
		\end{tabular}%
	}
\end{table}

\section{Continual Learning Method Extensions}
\label{Methods}
\subsection{Building Options}
Unlike standard image classification networks, the CLIP model comprises an image encoder and a text encoder. The dual encoder structure gives additional flexibility in designing continual learning algorithms. One could choose to update only the image encoder, or only the text encoder, or both of them, named as {\bf Whole Model (WM)}, {\bf Image Only (IO)}, and {\bf Text Only (TO)}. In the following parts, we first discuss the extension of existing continual learning approaches for CL-CLIP and then describe a simple-but-effective strategy proposed in this work.

\subsection{Extensions for CL-CLIP}
\label{extension}
In this section, we consider the extension of four existing methods for the CL-CLIP problem,  including LwF \cite{LwF}, GeoDL \cite{GeoDL}, IMM \cite{IMM} and RKR \cite{RKR}, where LwF and GeoDL (we only use the distillation in GeoDL, described below) represent the distillation-based approach, IMM is an adaptive-plasticity-based approach, and RKR is an architecture-based approach. 

\paragraph{LwF \cite{LwF}} LwF is a classic distillation-based continual learning method. LwF adds an additional distillation loss to encourage the posterior probabilities predicted from the updated model to be similar to the previous model. Specifically, it employs the following loss term:
\begin{align}
	&{\cal L} = {{\cal L}_{CE}} + \beta \cdot {{\cal L}_{LwF}}, \\
	&{{\cal L}_{LwF}}\left( {\mathbf{p}_{new}},{\mathbf{p}_{old}} \right) =  - \sum\limits_{c = 1}^{{K_p}} {p_{new}^c \cdot \log \left( {p_{old}^c} \right)} , \label{lwf}
\end{align}
where ${\mathbf{p}_{new}}$ and $\mathbf{p}_{old}$ are posterior probabilities estimated from the updated model and the original model. $p_{*}^c$ is the probability of class $c$ and $K_p$ denotes the number of previous classes. $\beta$ is the trade-off weight between the two losses.

\textit{CL-CLIP extension}: During training, the images and class prompts are encoded by previous and updated CLIP respectively, estimating ${\mathbf{p}_{new}}$ and $\mathbf{p}_{old}$ via \Cref{clip_prob}. It is notable that there is no explicit ``class" for original CLIP, we have to perform distillation on updated classes. For one-session training, the total 80 classes are trained with classification and distillation simultaneously. For multi-session training, the classification number is 10 because training data is limited on a session. But the distilled classes are both current classes and previous classes.

\begin{figure*}[t]
	\centering
	\includegraphics[width=0.8\textwidth]{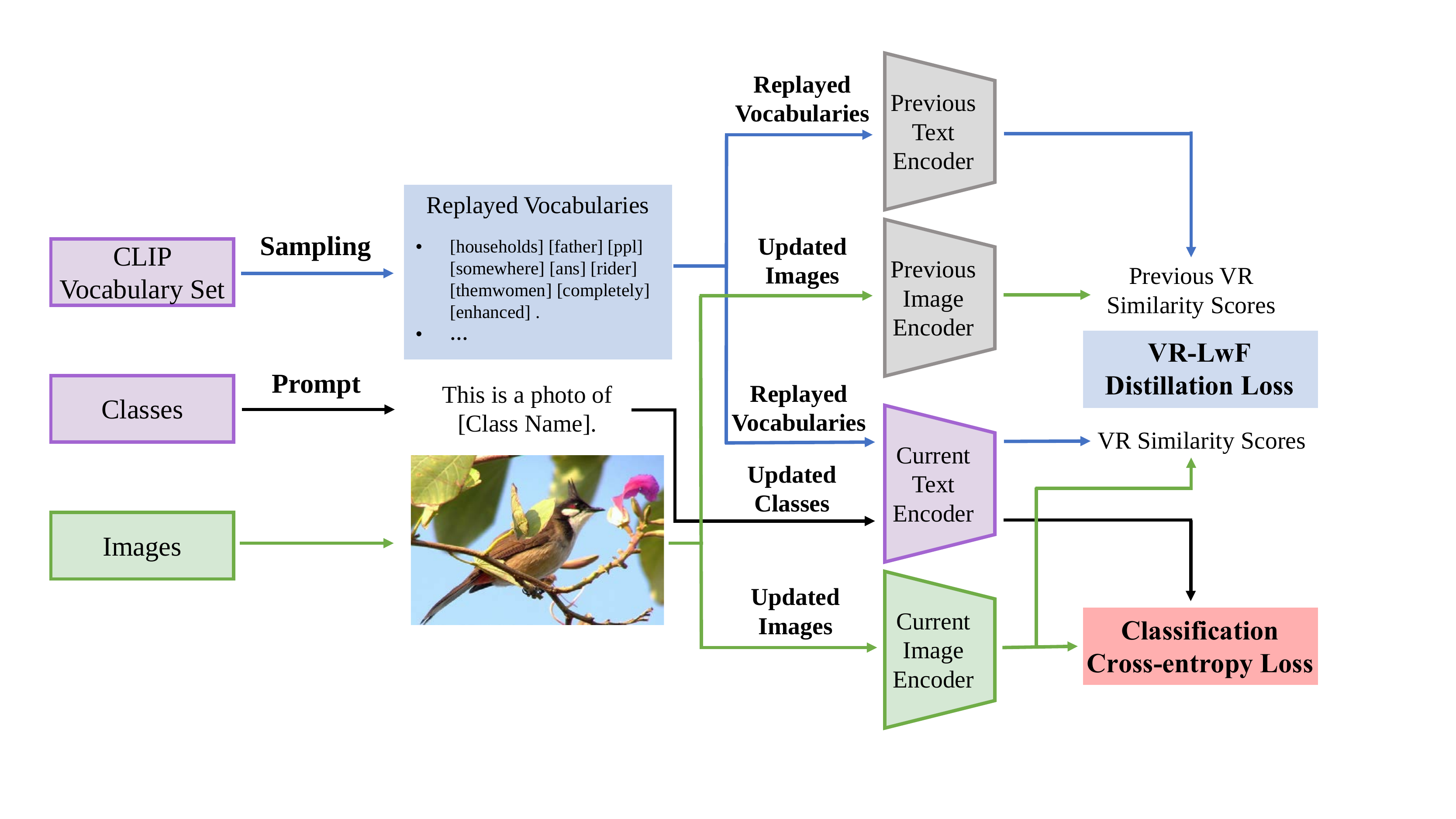}
	\caption{The framework of the proposed VR-LwF method. The pseudo vocabulary sequences are fed into both previous and current text encoders. The logits of the replayed classes are enforced. The model is trained with the combination of cross-entropy classification loss and replaying distillation loss. The grey part in the figure indicates the previous model and its outputs.}
	\label{VRLWF}
\end{figure*}

\paragraph{GeoDL \cite{GeoDL}} GeoDL is recently proposed knowledge distillation-based CL method. It first constructs low-dimensional manifolds for the features extracted from the previous and current models. Then GeoDL minimizes the representation dissimilarity by the geodesic similarity rather than the Euclidean similarity. We denote feature ${\bf{z}} \in \left\{ {{\bf{t}},{\bf{f}}} \right\}$ as CLIP visual or textual feature, ${\bf{z}}_{old}$ and ${\bf{z}}_{new}$ are from the old and new model, respectively. The feature matrices (${\bf{Z}}_{old}$ and ${\bf{Z}}_{new}$), which is usually stacked by visual or textual features in a batch, are decomposed to obtain subspaces ${\bf{P}}_{old}$ and ${\bf{P}}_{new}$. After the estimation of geodesic flow ${\bf{\Pi }}$ from the subspaces, the inner product of ${\bf{z}}_{old}$ and ${\bf{z}}_{new}$ are computed along ${\bf{\Pi }}$:

\begin{align}
	&{\bf{\Pi }}\left( \nu  \right) = \left[ {\begin{array}{*{20}{c}}
			{{{\bf{P}}_{old}}}&{\bf{R}}
	\end{array}} \right]\left[ {\begin{array}{*{20}{c}}
			{{{\bf{U}}_1}{\bf{\Gamma }}\left( \nu  \right)}\\
			{ - {{\bf{U}}_2}{\bf{\Sigma }}\left( \nu  \right)}
	\end{array}} \right],\\
	&{\bf{Q}} = \int_0^1 {{\bf{\Pi }}\left( \nu  \right){\bf{\Pi }}{{\left( \nu  \right)}^ \top }} d\nu , \\
	&{{\mathop{\rm g}\nolimits} _\Pi }\left( {{{\bf{z}}_{old}},{{\bf{z}}_{new}}} \right) = {{\bf{z}}_{new}}^ \top {\bf{Q}}{{\bf{z}}_{old}},
\end{align}
where ${\bf{R}}$ is orthogonal complement of ${\bf{P}}_{old}$, ${\bf{\Gamma }\left( \nu  \right)}$ and ${\bf{\Sigma }\left( \nu  \right)}$ are two diagonal matrices. ${{\mathop{\rm g}\nolimits} _\Pi }\left( \cdot , \cdot \right)$ is the inner product, ${\bf{Q}}$ is the matrix which defines the manifold structure between features of two phases. Details and proofs can be found in its original paper \cite{GeoDL}.

\textit{CL-CLIP extension}:
The original GeoDL method is conducted with replaying samples, which is not applicable for CL-CLIP. Thus, we only treat it as a distillation method and perform it for both the text and image embeddings. During training, the images and prompts are inputted to old model and new model, extracting feature matrices, ${\bf{F}}_{old}, {\bf{F}}_{new}$ and ${\bf{T}}_{old}, {\bf{T}}_{new}$, for vision and text respectively. Notice that for MST, the text feature matrix only contains current and previous classes, which is the same as LwF. Then the subspaces and manifolds ${\bf{Q}}^i, {\bf{Q}}^t$ are estimated, where ${\bf{Q}}^i$ is manifold between old and new image features and ${\bf{Q}}^t$ is for text. The GeoDL distillation aims to minimize the inner product of image and text features between old and new model:

\begin{align}
	&{{\cal L}_{GeoDL}^i} = - \frac{{{{\bf{f}}_{new}}^ \top {{\bf{Q}}^i}{{\bf{f}}_{old}}}}{{\left\| {{{{\bf{Q}}^i}^{1/2}}{{\bf{f}}_{new}}} \right\|\left\| {{{{\bf{Q}}^i}^{1/2}}{{\bf{f}}_{old}}} \right\|}}, \\
	&{{\cal L}_{GeoDL}^t} = - \frac{{{{\bf{t}}_{new}^c}^ \top {{\bf{Q}}^t}{{\bf{t}}_{old}^c}}}{{\left\| {{{{\bf{Q}}^t}^{1/2}}{{\bf{t}}_{new}^c}} \right\|\left\| {{{{\bf{Q}}^t}^{1/2}}{{\bf{t}}_{old}^c}} \right\|}},
\end{align}
where ${\bf{f}}$ is visual embedding and ${\bf{t}}^c$ is textual embedding of class $c$. The distillation losses are combined with classification loss by ${\cal L} = {{\cal L}_{CE}} + \beta \cdot {{\cal L}_{GeoDL}^i} + \beta \cdot {{\cal L}_{GeoDL}^t}$.

\paragraph{IMM \cite{IMM}} IMM is a adaptive-plasticity-based continual learning method. There are two variants in IMM (\emph{i.e.}, IMM-Mean and IMM-Mode). IMM-Mode combines two networks with the Laplacian approximation \cite{Laplacian}, but requires the training information of previous phases, which is expensive for the CL-CLIP model. IMM-Mean simply averages the parameters of current and preceding models, which suits CL-CLIP well. The IMM-mean approach is also employed in a recent work called WiSE-FT \cite{WiSE-FT} to benefit out-of-distribution generalization of CLIP. In WiSE-FT \cite{WiSE-FT}, the IMM-mean is slightly modified to leverage the weighted-combination rather than a direct average. Please note that WiSE-FT \cite{WiSE-FT} and our method are targeting different problems, \emph{i.e.}, they try to improve CLIP's domain generalization capability while we are targeting the CL-CLIP problem. 

\textit{CL-CLIP extension}: IMM can be applied to the text-encoder and/or image encoder. In all cases, we generate the final model by taking the weighted average of fine-tuned CLIP and the original CLIP:
\begin{align}
	&{\bf{f}} = {{\mathop{\rm CLIP}\nolimits} _i}\left( {{\bf{x}};\left( {1 - \alpha } \right) \cdot {{\bf{\theta }}_{i,old}} + \alpha  \cdot {{{\bf{\theta}}}_{i,new}}} \right), \\
	&{{\bf{t}}^c} = {{\mathop{\rm CLIP}\nolimits} _t}\left( {{{\bf{y}}^c};\left( {1 - \alpha } \right) \cdot {{\bf{\theta }}_{t,old}} + \alpha  \cdot {{{\bf{\theta}}}_{t,new}}} \right), 
\end{align}
where ${{\bf{\theta }}_{*,old}}$ and ${{\bf{\theta }}_{*,new}}$ are model parameters before and after training. Depending on the updating options, we could have three variants of CL-CLIP IMM. During training, the old model is original CLIP for OST but the previous session model for MST.

\paragraph{RKR \cite{RKR}} 
RKR keeps the previously trained parameters fixed and adds very few extra learnable parameters. There are two modules in RKR, including the rectification generator (RG) and the scaling factor generator (SFG). RKR transforms the fixed model as follows:
\begin{align}
	&{{\bf{\theta }}_w} = {{\bf{\theta }}_w} + {{\bf{r}}_w},\\
	&{{\bf{o}}_w} = {{\bf{o}}_w} \cdot {{\bf{s}}_w},
\end{align}
where ${{\bf{\theta }}_w}$ is the frozen parameter for layer $w$, ${{\bf{r}}_w}$ is generated weight rectification, ${{\bf{o}}_w}$ is the output of layer and it is scaled by learned ${{\bf{s}}_w}$.

\textit{CL-CLIP extension}: Both the text encoder and image encoder of the CLIP are transformer \cite{Transformer} architectures. We only add RG to linear layers in the transformer, but according to our experiments, the RG causes the severe forgetting of zero-shot ability. So we only apply SFG to each block's output and ${{\bf{s}}_w}$ are initialized by all one vectors. During both OST and MST training, the only fine-tuned part is the added parameters. Notice that, in MST, the RKR parameters are installed at the first session and continually trained in all tasks.

\section{Method} 
\label{VRD}
One of the most successful families of continual learning algorithms is the replaying-based approach. Specifically, it stores a few samples from each previously learned class in a replaying memory. The replaying method tries to maintain their corresponding outputs from the old model when training on a new task. Unfortunately, for the CL-CLIP problem, the replaying method is not suitable for two reasons: (1) In the CLIP, there is no explicit ``class'' concept. (2) CLIP covers a huge number of visual concepts, which is memory-consuming to store the exemplars.  

However, we recognize that the replying scheme could be applied to the text encoder, which creates a solution for CL-CLIP. The idea is motivated by the fact that the classifier of classes is derived from the text encoder by prompted sentences, the text encoder is a mapping function from class names to classifier weights (see \Cref{text_encode}). The input of this mapping function is essentially the combination of tokens from a vocabulary set, which is available without the need for images. Thus, we can generate sentences from the vocabularies and perform distillation -- this allows us to simulate the zero-shot and previous classes during training, alleviating the catastrophic forgetting. To be specific, the replayed vocabularies can be regarded as pseudo-classes, we implement distillation on their logits. This is similar to the LwF method, but LwF is executed on true classes while our method utilizes replayed classes. Hence, the method is entitled Learning without Forgetting via Replayed Vocabularies (VR-LwF).

\begin{algorithm}[t]
	\centering
	\caption{Replayed Vocabularies Building.}
	\label{VRALG}
	\begin{algorithmic}[1]
		\renewcommand{\algorithmicrequire}{\textbf{Input:}}
		\REQUIRE
		CLIP vocabulary set $V_{CLIP}$, sampling length $M$, and replaying class number $K_s$.
		
		\renewcommand{\algorithmicrequire}{\textbf{Output:}}
		\REQUIRE 
		A set $\mathbf{Y}^s$ of $K_s$ elements, which contains replayed classes $\mathbf{y}^s$.
		\STATE Initialize $\mathbf{Y}^s$ with an empty set.
		\FOR {$s = 1$ to $K_s$}
		\STATE Set $\mathbf{y}^s$ as an empty sequence. 
		\FOR {$i = 1$ to $M$}
		\STATE $w \leftarrow$ Random select a vocabulary in $V_{CLIP}$;
		\STATE $\mathbf{y}^s \leftarrow$ Add the word $w$ into sequence;
		\ENDFOR
		\STATE $\mathbf{Y}^s \leftarrow$ Add the replayed vocabularies $\mathbf{y}^s$ into set.
		\ENDFOR
	\end{algorithmic}
\end{algorithm}

\begin{table}[t]
	\centering
	\small
	\caption{Retrieval recalls (\%) of Original CLIP model when random shuffle the input text.}
	\label{Shuffle}
	\resizebox{0.45\textwidth}{!}{%
		\begin{tabular}{c|cccccc}
			\hline
			Method & TR@1 & TR@5 & TR@10 & IR@1 & IR@5 & IR@10 \\
			\hline 	
			VSE++ \cite{VSE} & 58.30 & - & 93.30 & 43.60 & - & 87.80 \\ 
			
			Non-shuffle & 69.34 & 91.08 & 95.70 & 49.67 & 79.17 & 88.77 \\	
			
			Shuffle & 54.06 & 81.38 & 90.26 & 35.64 & 66.00 & 78.72 \\ \hline
			
		\end{tabular}%
	}
\end{table}

A problem for this idea is how can we generate sentences from the existing vocabularies, e.g., should we follow certain grammar structure? In our work, we take a crude approximation by completely ignoring the grammar. This scheme is motivated by the following observation: we observe that by randomly shuffling the word order of a sentence, the text-to-image retrieval capability of CLIP can be largely preserved. As seen from \Cref{Shuffle}, by shuffling the word order, the TR@$n$ of the retrieval performance can still be quite reasonable and can even be competitive to an existing retrieval approach VSE++ \cite{VSE}. This inspires us to generate (pseudo-)sentences by directly stacking randomly chosen tokens from the vocabulary. Formally, we arbitrarily sample $M$ words from the CLIP vocabulary to constitute a text input $\mathbf{y}^s$, as shown in \Cref{VRALG}. The pseudo-sentence is fed into both the current text encoder and the original text encoder to obtain the embeddings ${\bf{t}}_{new}^{s}$ and ${\bf{t}}_{old}^{s}$. Meanwhile, the image embeddings of current model and previous model are ${\bf{f}}_{new}$ and ${\bf{f}}_{old}$. Then a distillation loss is applied:

\begin{align}
	&p_{old}^s = \frac{{\exp \left( {{\tau _{old}} \cdot \left\langle {{{\bf{f}}_{old}},{\bf{t}}_{old}^s} \right\rangle } \right)}}{{\sum\nolimits_{j = 1}^{{K_s}} {\exp \left( {{\tau _{old}} \cdot \left\langle {{{\bf{f}}_{old}},{\bf{t}}_{old}^j} \right\rangle } \right)} }},\\
	&p_{new}^s = \frac{{\exp \left( {{\tau _{new}} \cdot \left\langle {{{\bf{f}}_{new}},{\bf{t}}_{new}^s} \right\rangle } \right)}}{{\sum\nolimits_{j = 1}^{{K_s}} {\exp \left( {{\tau _{new}} \cdot \left\langle {{{\bf{f}}_{new}},{\bf{t}}_{new}^j} \right\rangle } \right)} }},\\
	&{\cal L}_{LwF}^{VR}\left( {{\bf{p}}_{new}^s,{\bf{p}}_{old}^s} \right) =  - \sum\limits_{s = 1}^{{K_s}} {p_{new}^s \cdot \log \left( {p_{old}^s} \right)},
\end{align}
where ${\mathbf{p}_{new}^s}$ and $\mathbf{p}_{old}^s$ are probability distribution among the pseudo classes. $p_{*}^s$ is the probability of class $s$ and $K_s$ is the number of replayed classes. Combining with the cross-entropy loss for training the updated classes, the final loss function is given by ${\cal L} = {{\cal L}_{CE}} + \beta \cdot {{\cal L}_{LwF}^{VR}}$. The illustration is shown in \Cref{VRLWF}. The replayed vocabularies act as unseen and previous classes, but for MST, it is a more effective way to directly append former classes to the replayed vocabularies. The loss function would be ${\cal L} = {{\cal L}_{CE}} + \beta \cdot {\left( {{\cal L}_{LwF}^{VR}} + {{\cal L}_{LwF}^{Pre}} \right)}$, where ${{\cal L}_{LwF}^{Pre}}$ is a LwF loss on old session classes. Note that this is different with LwF in \Cref{lwf}, the LwF extension distils on all seen classes (including previous classes and current classes), while ours only contains previous classes. It is not a good choice for LwF extension to use only previous classes, since it will degenerate to direct fine-tuning at the first session, but VR-LwF can still use other replayed classes.

\section{Experiments}
\label{experiment}
This section presents our experimental studies on the CL-CLIP problem. We started by introducing the implementation details and evaluation metrics. Then, we conduct a series of experiments to answer the following questions: (1) Whether the CLIP model suffers from the catastrophic forgetting issue? (2) Can current continual learning methods, at least the four selected representative methods, be extended to alleviate the forgetting issue? (3) What experiences are explored under the two training settings? (4) Can the proposed VR-LwF method lead to improved performance? If yes, what is the impact of the hyper-parameters in the proposed method?

\subsection{Implementation Details and Evaluation Metrics}

\paragraph{Training Details} As mentioned above, we evaluate CL-CLIP under two settings: one-session training (OST) and multi-session training (MST). We fine-tune 15 epochs for OST, the initial learning rate is 1e-6 and it will decay by 0.1 after 10 epochs. The number of epochs for each MST session is 10, where are 80 epochs in total. Though this is much more than OST, the training step is only equivalent to 10-epoch OST. The learning rate of MST is 1e-6 for each session, without decline. All compared methods are optimized with Adam \cite{Adam} optimizer, and we set the batch size as 100. For our VR-LwF, the length of pseudo-sentence is 10, and the number of pseudo-sentences is 100 per batch.

\begin{table*}[t]
	\centering
	\scriptsize
	\caption{Results (\%) of One-Session Training. The Original CLIP model does not participate in the comparison. The best scores are marked in bold. Note that the best results in each group are underlined.}
	\label{OST}
	\tiny
	\resizebox{0.85\textwidth}{!}{%
		\begin{tabular}{c|cccccc|c}
			\hline
			Method & UT-Acc & ZS-Acc & TR@1 & TR@5 & IR@1 & IR@5 & A-Acc \\
			\hline 
			
			Original CLIP & 25.36 & 62.62  & 69.34 & 91.08 & 49.67 & 79.17 & 43.99 \\ \hline	
			
			FT-WM & 74.21 & 44.17  & 32.54 & 59.50 & 25.49 & 53.55 & 59.19 \\
			LwF-WM & \underline{\textbf{74.41}} & 58.75  & 62.34 & 86.34 & 45.86 & 76.1 & 66.58 \\ 
			GeoDL-WM & 74.30 & 60.50 & 65.18 & 87.54 & 46.37 & 76.14 & 67.40 \\ 
			IMM-WM & 71.25 & 59.94 & 60.66 & 85.08 & 43.03 & 73.60 & 65.60 \\ 
			RKR-WM & 56.60 & \underline{63.82} &  \underline{69.36} & \underline{\textbf{90.76}} & \underline{\textbf{51.27}} & \underline{\textbf{80.60}} & 60.21 \\
			\textbf{VR-LwF-WM} & 72.82 & 62.03 & 68.34 & 89.96 & 47.82 & 77.41 & \underline{\textbf{67.43}}\\ 
			\hline		
			
			FT-IO & 74.11 & 38.84  & 34.68 & 61.30 & 23.77 & 50.83 & 56.48 \\
			LwF-IO & 73.91 & 57.19  & 62.70 & 86.72 & 48.27 & 78.30 & 65.55 \\ 
			GeoDL-IO & 74.23 & 57.03 & 62.48 & 86.74 & 47.15 & 76.85 & 65.63 \\ 
			IMM-IO & 70.25 & 57.21  & 61.74 & 86.10 & 43.29 & 73.02 & 63.73 \\ 
			RKR-IO & 50.11 & \underline{62.26} & \underline{69.96} & \underline{90.66} & \underline{51.04} & \underline{80.37} & 56.19\\
			\textbf{VR-LwF-IO} & \underline{74.69} & 58.58 & 64.76 & 88.42 & 46.97 & 76.77 & \underline{66.63}\\ 
			\hline
			
			FT-TO & \underline{64.86} & 56.44  & 61.26 & 84.64 & 41.16 & 70.65  & 60.65 \\
			LwF-TO & 64.35 & 59.52  & 66.16 & 88.54 & 45.04 & 74.85 & 61.93\\ 
			GeoDL-TO & 64.52 & 60.50 & 66.00 & 88.52 & 45.30 & 75.79 & 62.51 \\ 
			IMM-TO & 63.50 & 61.31 & 68.50 & 89.50 & 47.43 & 77.04 & 62.41\\	
			RKR-TO & 54.18 & \underline{\textbf{64.74}} & \underline{\textbf{69.48}} & \underline{90.66} & \underline{50.11} & \underline{79.58} & 59.46\\ 
			\textbf{VR-LwF-TO} & 64.85 & 61.79 & 68.52 & 90.48 & 49.41 & 78.90 & \underline{63.32}\\ 
			\hline
			
		\end{tabular}%
	}
\end{table*}

\paragraph{Evaluation Metrics} We use the following metric to evaluate the performance of the updated model with different CL-CLIP methods:

\begin{itemize}
	\item UT-Acc: the model accuracy of the updated task test set. We expect a high UT-Acc when the model has successfully fine-tuned from the update-task training data. Please note that in OST, the accuracy is overall accuracy (calculated by the percentage of the correct prediction versus the total number of samples). For the MST setting, UT-Acc is calculated by averaging over all sessions. Due to the number of samples being different for each session in MST, the UT-Acc of MST is not directly comparable with that of OST even for the same set of predictions (25.36\% and 39.80\% in \Cref{OST} and \Cref{MST}). 
	
	\item ZS-Acc: the zero-shot learning test accuracy. Generally speaking, there is a trade-off between UT-Acc and ZS-Acc, a model that achieves high UT-Acc may come with sacrificing its zero-shot learning capability and result in a low ZS-Acc.
	
	\item R@$k$: Recall at 1 and 5 (TR@1, TR@5 for text retrieval and IR@1, IR@5 for image retrieval) are used to evaluate the image-text and text-image matching capability.
	
	\item Bwt: backward transfer is only conducted for MST. It is the average drop of accuracy on previous sessions after fine-tuning with the current task, which is defined in \cite{GEM}: ${\mathop{\rm Bwt}\nolimits}  = \frac{1}{{S - 1}}\sum\nolimits_{i = 1}^{S - 1} {Ac{c_{S,i}} - Ac{c_{i,i}}}$, where $S$ is the current session number, $Acc_{S,i}$ is the accuracy of session $i$ classes after current fine-tuning and $Acc_{i,i}$ is the accuracy of task $i$ after the $i$-th session training. As the forgetting often appears in continual learning, Bwt is usually small than 0, and the smaller the Bwt, the worse the forgetting.
	
	\item A-Acc: the average of UT-Acc and ZS-Acc. Since there is a trade-off between UT-Acc and ZS-Acc, it will be less convenient to judge the performance of an algorithm by examining two accuracies. We thus use their average accuracy as a single indicator for the goodness of CL-CLIP algorithms.
\end{itemize}

\paragraph{Hyper-parameter Choices} Hyper-parameters of each method are searched from the validation set. Specifically, we choose the hyper-parameter that brings the highest A-Acc on the validation set. The hyper-parameters to be optimized are the combination weight $\alpha$ in IMM, and the trade-off weights $\beta$ for the distillation terms in LwF, GeoDL, VR-LwF.

\paragraph{Compared Methods} As discussed in \Cref{Methods}, we can extend an existing CL method with three options, that is, updating the whole model (WM), only the image encoder (IO), or only the text encoder (TO). We apply those three options to both the fine-tuning baseline (FT) and the four CL methods, \emph{i.e.}, LwF, GeoDL, IMM, and RKR. We use the notation ``method-WM/IO/TO'' to denote these variants, including our VR-LwF.

\begin{table*}[t]
	\centering
	\caption{Results (\%) of Multi-Session Training. The notation is the same as the one-session training. Joint-FT model reveals the upper bound of UT-Acc, because of its serious damage on zero-shot performance, other metrics are not reported.}
	\label{MST}		
	\scriptsize
	\resizebox{0.9\textwidth}{!}{%
		\begin{tabular}{c|ccccccc|c}
			\hline
			Method & UT-Acc & Bwt & ZS-Acc & TR@1 & TR@5 & IR@1 & IR@5 & A-Acc \\
			\hline
			Original CLIP & 39.80 & - & 62.62 & 69.34 & 91.08 & 49.67 & 79.17 & 51.21 \\ 
			Joint-FT & 67.33 & - & - & - & - & - & - & - \\ \hline				
			
			FT-WM & 51.51 & -18.45 & 50.01  & 46.40 & 73.82 & 35.94 & 66.56 & 50.76\\
			LwF-WM & 52.65 & -1.60 & 58.43  & 63.72 & 87.44 & 44.22 & 73.81 & 55.54 \\ 
			GeoDL-WM & 53.27 & -11.14 & 60.09 & 63.60 & 87.14 & 42.81 & 73.66 & 56.68  \\ 
			IMM-WM & 54.75 & -2.69 & 61.91 & 65.74 & 88.90 & 48.34 & 77.80 & 58.33\\ 
			RKR-WM & 45.37 & \underline{0.68} & \underline{64.00} & \underline{\textbf{69.78}} & \underline{\textbf{90.96}} & \underline{\textbf{51.37}} & \underline{\textbf{80.80}} & 54.68 \\
			\textbf{VR-LwF-WM}  & \underline{\textbf{58.03}} & -1.36 & 63.91 & 68.12 & 90.38 & 48.61 & 77.96 & \underline{\textbf{60.97}}\\ 
			\hline
			
			FT-IO & 49.05 & -22.23 & 47.53 & 49.26 & 75.86 & 36.17 & 66.56 & 48.29\\
			LwF-IO & 48.75 & 1.50 & 59.58 & 65.12 & 88.42 & 48.89 & 78.39 & 54.16 \\ 
			GeoDL-IO & 54.90 & -4.47 & 59.48  & 65.46 & 88.88 & 48.80 & 78.19 & 57.19\\ 
			IMM-IO & 51.33 & -2.94 & 60.23 & 65.90 & 88.60 & 48.03 & 77.80 & 55.78\\	
			RKR-IO & 43.68 & \underline{0.25} & \underline{61.99} & \underline{69.82} & \underline{90.78} & \underline{51.26} & \underline{80.60} & 52.83\\ 
			\textbf{VR-LwF-WM}  & \underline{56.87} & -2.76 & 59.09 & 64.32 & 88.18 & 48.18 & 78.02 & \underline{57.98}\\	
			\hline
			
			FT-TO & 52.27 & -3.51 & 55.45 & 62.26 & 87.20 & 41.54 & 71.86 & 53.86\\
			LwF-TO & 51.27 & -1.12 & 59.77 & 68.68 & 90.10 & 42.62 & 73.27 & 55.52\\ 
			GeoDL-TO & 52.64 & -6.09 & 55.84 & 63.26 & 87.28 & 37.54 & 68.67 & 54.24\\ 
			IMM-TO & 51.03 & -0.30 & 61.61 & 67.74 & 90.28 & 48.31 & 77.98 & 56.32\\ 
			RKR-TO & 44.24 & \underline{\textbf{1.64}} & \underline{\textbf{64.89}} & \underline{69.24} & \underline{90.82} & \underline{49.88} & \underline{79.47} & 54.57\\  			
			\textbf{VR-LwF-WM}  & \underline{53.80} & -0.24 & 62.57 & 68.94 & 90.52 & 48.87 & 78.61 & \underline{58.18}\\ 
			\hline
			
		\end{tabular}%
	}
\end{table*}

\subsection{Experimental Results}

\begin{figure*}[t]
	\centering
	\includegraphics[width=0.8\textwidth]{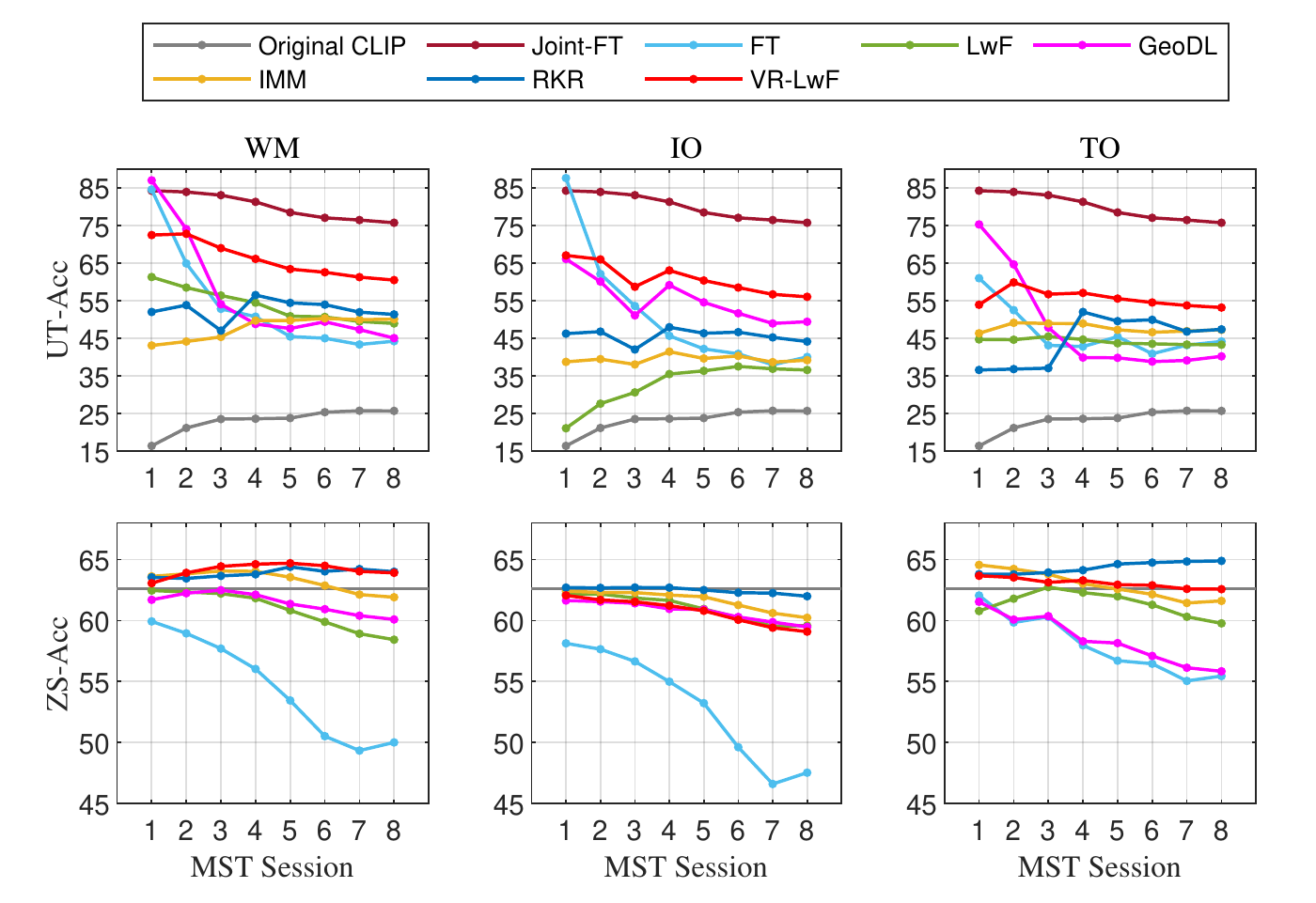}
	\caption{Accuracy (\%) of methods in each MST session. The first row of charts is updated accuracy, the second row is for ZS-Acc. From left to right, each column is for WM, IO, and TO variants in sequence. The Original CLIP and Joint-FT are lower and upper bounds of UT-Acc, respectively.}
	\label{Curve}
\end{figure*}

In this section, we compare different CL-CLIP approaches. \Cref{OST} presents an overview of one-session training results. \Cref{MST} presents the results of MST, which reports after the training of the last session. We also introduce the accuracy curves in \Cref{Curve}, note that the UT-Acc in the figure is overall accuracy rather than average accuracy.\footnote{This is same as continual learning literature, the average accuracy and Bwt are usually reported in the table, like \Cref{MST}, and overall accuracy is for curves.} We summarize our findings as follows:

\paragraph{CLIP suffers from catastrophic forgetting}
We firstly focus on FT, \emph{i.e.}, directly fine-tuning the model without applying any continue learning algorithms. From the results in \Cref{OST}. We can find that fine-tuning leads to significantly improved performance on the updated task as expected. However, it causes damage to the zero-shot learning and image-text matching capabilities. We observe a sharp performance drop of ZS-Acc, TR@$k$, and IR@$k$ for either WM, IO, or TO variants. This result suggests that the CLIP model suffers from catastrophic forgetting. Likewise, this can be seen in \Cref{MST}, the MST setting. The small Bwt also demonstrates that the previously learned classes are forgotten after new session training.

\paragraph{Different updating options perform dissimilarly}\label{FT_opt}
As observed from \Cref{OST} and \Cref{MST}, the choice of the to-be-updated module would significantly impact the performance. Generally, \textbf{the TO variant is slightly safer} than the other two, but the accuracy of updated task is limited. Taking FT-TO as an example, its ZS-Acc only drops from 62.62\% to 56.44\%/55.45\% (OST/MST), which is the best among FT variants. But the UT-Acc is 64.86\% (OST, since the UT-Acc of MST is also influenced by forgetting, its UT-Acc is higher than WM and IO), lower than FT-IO and FT-WM (74.21\%/74.11\%). On the contrary, \textbf{IO variant suffers from the worst forgetting} among all variants, the ZS-Acc of almost all methods is lower than WM and TO. A possible explanation for this might be that the gradient of TO fine-tuning loss solely propagates to the discrete tokens (the prompt and class names) in the vocabulary set, so it does not impact other vocabularies. However, the performance still declines because model parameter change brings the mismatch with these vocabularies. But texts are more discrete than images, which means there would be more space for the shift of text embeddings, hence the performance loss of TO variant is lower than IO. What is interesting about the retrieval metrics is that the TR@$k$ of IO and IR@$k$ of TO usually decline more. Finally for the WM variant, only fine-tuning is not the best choice (A-Acc of FT-WM is worse than FT-TO's). \textbf{But the WM variant performs best by cooperating with CL methods.} This means that when fine-tuning text encoder and image encoder simultaneously, the change of two parts can be slighter than forcing only one part to update, with the help of CL methods. This also indicates that it is able to get better performance by fine-tuning more parameters.

\paragraph{There are CL methods to alleviate the forgetting issue, but only for OST}
Closer inspection of the \Cref{OST} shows GeoDL is the best CL method, whose performance is better than others except for our VR-LwF. GeoDL-WM even gets the superior A-Acc, which is close to ours (67.40\% vs. 67.43\%). But under MST setting, it is incapable for forgetting challenge, the gap with VR-LwF is blown-up and it is worse than IMM. This can be inferred from the ZS-Acc drop in OST. ZS-Acc of GeoDL-WM decreases around 2\% (60.50\% vs. 62.62\%), comparing with Original CLIP. This forgetting will be further distinguished when training tasks sequentially. Previous classes will become zero-shot classes in current session, which means they prone to be forgotten, resulting in the inferior UT-Acc. It can be observed that the ZS-Acc of GeoDL-WM is near between OST and MST, so the gap is mainly caused by UT-Acc. It also happens on the second-best CL method (in OST) LwF. This illustrates that existing CL methods may alleviate the forgetting in one session training but they are weak for multi-session CL-CLIP.

\paragraph{CL-CLIP is different from traditional continual learning}
In \Cref{Curve} we plot the performance curves of compared methods. For updated task, We introduce Original CLIP and Joint-FT as the lower and upper-performance bounds. The Joint-FT is built, as same as CL literature, by fine-tuning with all seen data. Its average accuracy of session 8 is the same as it in \Cref{MST}. In the bottom charts, the grey line presents Original CLIP ZS-Acc (62.62\%). There are some interesting findings. First of all, some UT-Acc curves even increase, such as LwF, IMM, and RKR, which is strange in continual learning. The main reason lies in the paradigm of CLIP, that the classifier weights are not independent. As introduced above, the text encoder (classifier) of CLIP is a mapping function, so the change of this mapping would bring changes to all classes, this is different from the fully-connected classifier whose weights are independent. \textbf{The rising curve means the subsequent training improves the former classification.} However, this property also leads to a problem that, \textbf{to keep the zero-shot capability, CL methods limit the performance of updated classes}. The CL updating is effective partly, as we can see that all methods beat the Original CLIP for UT-Acc. But compared with FT, their gain of ZS-Acc is much higher than UT-Acc. And some CL methods are limiting on updated task. This is apparent on IMM, it is the third-best method on ZS-Acc, while its UT-Acc is low-rank among CL methods. This finding is consistent with the above discussion that the loss of CL methods on MST is mainly caused by the drop of UT-Acc. The proposed VR-LwF is superior to CL extensions. The ZS-Acc is equal to the Original CLIP and UT-Acc is the best. Unfortunately, VR-LwF is still around 15\% lower than the Joint-FT. A further study with more focus on increasing the performance of updated classes is therefore suggested.

\paragraph{RKR keeps the original performance best, but learns least}
Surprisingly, the RKR method was found to get the best zero-shot performance. The ZS-Acc and retrieval recalls are all the best in each variant group and each training setting (underlined in \Cref{OST} and \Cref{MST} means the best in each group). Some of them are even higher than the original CLIP model, such as ZS-Acc and IR. These results are related to RKR which keeps the CLIP model frozen and trains additional parameters. Furthermore, it should be noted that RKR's updated task performance is the worst among all methods, usually 10\% lower than others. The major factor is also the adding parameters, as mentioned before, the RKR extension only contains the scaling factor part. This is lacking compared with other fine-tuning methods, but adding the rectification part makes the CLIP forget all zero-shot abilities.

\begin{table}[t]
	\centering
	\small
	\caption{Results (\%) of one-session VR-LwF-WM with different replaying methods. We examine the influence of different random-sampling lengths and the caption replaying.}
	\label{M_hyper}
	\tiny
	\resizebox{0.48\textwidth}{!}{%
		\begin{tabular}{c|cccc|c}
			\hline
			Method & UT-Acc & ZS-Acc & TR@1 & IR@1 & A-Acc \\
			\hline 
			
			Original CLIP & 25.36 & 62.62  & 69.34 & 49.67 & 43.99 \\ \hline	
			
			$M$=10 & 72.82 & 62.03 & 68.34 & \textbf{47.82} & 67.43 \\
			$M$=20 & 74.11 & 61.06 & 66.56 & 47.14 & 67.59 \\
			$M$=30 & 74.41 & 60.87 & 65.80 & 46.47 & \textbf{67.64} \\	
			$M$=50 & \textbf{74.45} & 60.75 & 64.90 & 46.09 & 67.60 \\
			$M$=70 & 74.26 & 60.69 & 66.22 & 46.83 & 67.47 \\ \hline
			Caption Corpus & 71.57 & \textbf{62.52} & \textbf{69.04} & 47.36 & 67.05 \\
			
			\hline
			
		\end{tabular}%
	}
\end{table}

\begin{table}[t]
	\centering
	\scriptsize
	\caption{Results (\%) of one-session VR-LwF-WM with different replaying numbers.}
	\label{Ks_hyper}
	\resizebox{0.48\textwidth}{!}{%
		\begin{tabular}{c|cccc|c}
			\hline
			Method & UT-Acc & ZS-Acc & TR@1 & IR@1 & A-Acc \\
			\hline 
			
			Original CLIP & 25.36 & 62.62  & 69.34 & 49.67 & 43.99 \\ \hline	
			
			$K_s$=50 & 73.19 & 61.67 & 67.64 & 47.28 & 67.43 \\
			$K_s$=100 & 72.82 & \textbf{62.03} & \textbf{68.34} & 47.82 & 67.43 \\
			$K_s$=200 & \textbf{73.09} & 61.86 & 67.76 & 47.98 & \textbf{67.47} \\	
			$K_s$=300 & 72.94 & 61.89 & 67.88 & 47.86 & 67.42 \\
			$K_s$=400 & 72.99 & 61.79 & 67.92 & \textbf{48.13} & 67.39 \\ 
			
			\hline
			
		\end{tabular}%
	}
\end{table}

\paragraph{The proposed VR-LwF achieves the overall best performance}
\Cref{OST} and \Cref{MST} are quite revealing in the superior performance of our proposed VR-LwF method. The best A-Acc of OST and MST both belong to VR-LwF-WM (67.43\% and 60.97\%). Also, for each VR-LwF variant, they all get the best A-Acc in their group. But the performance trend is consistent with former discussion, where TO is safer, IO is substandard, and WM is capable. For zero-shot abilities, their ZS-Acc and recalls are usually the second best, which is only below the RKR discussed before. The only difference between VR-LwF and LwF extension is to distill on which classes. For example in OST, LwF distills on updated classes and VR-LwF distills on replayed classes, comparing their outcomes, replaying is more advantageous. Though LwF is higher on UT-Acc (74.41\% vs. 72.82\%), VR-LwF's ZS-Acc is better (62.03\% vs. 58.75\%), it can be seen that the recalls of LwF are worse too. Moreover, as mentioned above, the gap is distinct in MST.

\begin{table}[t]
	\centering
	\small
	\caption{Results (\%) of multi-session VR-LwF-WM with different replaying methods. We examine the influence of different random-sampling lengths and the caption replaying.}
	\label{MM_hyper}
	\resizebox{0.49\textwidth}{!}{%
		\begin{tabular}{c|ccccc|c}
			\hline
			Method & UT-Acc & Bwt & ZS-Acc & TR@1 & IR@1 & A-Acc \\
			\hline 
			
			Original CLIP & 39.80 & - & 62.62  & 69.34 & 49.67 & 51.21 \\ \hline	
			
			$M$=10 & 58.03 & -1.36 & \textbf{63.91} & 68.12 & 48.61 & 60.97 \\
			$M$=20 & 58.58 & -1.13 & 63.61 & 68.00 & \textbf{48.62} & 61.10 \\
			$M$=30 & \textbf{58.60} & -1.73 & 63.66 & 68.32 & 48.31 & 61.13 \\	
			$M$=50 & 58.53 & -2.02 & 63.76 & 68.42 & 47.90 & 61.14 \\
			$M$=70 & 58.48 & -1.45 & 63.88 & \textbf{69.06} & 48.45 & \textbf{61.18} \\ \hline
			Caption Corpus & 57.30 & \textbf{-0.75} & 61.95 & 68.80 & 47.52 & 59.62 \\	
			\hline
			
		\end{tabular}%
	}
\end{table}

\begin{table}[t]
	\centering
	\small
	\caption{Results (\%) of multi-session VR-LwF-WM with different replaying numbers.}
	\label{MKs_hyper}
	\resizebox{0.48\textwidth}{!}{%
		\begin{tabular}{c|ccccc|c}
			\hline
			Method & UT-Acc & Bwt & ZS-Acc & TR@1 & IR@1 & A-Acc \\
			\hline 
			
			Original CLIP & 39.80 & - & 62.62  & 69.34 & 49.67 & 51.21 \\ \hline	
			
			$K_s$=50 & 58.05 & -1.17 & \textbf{63.98} & 68.46 & 48.70 & 61.02 \\
			$K_s$=100 & 58.03 & -1.36 & 63.91 & 68.12 & 48.61 & 60.97 \\
			$K_s$=200 & 58.12 & \textbf{-0.65} & 63.92 & 68.26 & 48.98 & 61.02 \\
			$K_s$=300 & \textbf{58.43} & -0.86 & 63.84 & \textbf{68.76} & 48.99 & \textbf{61.14} \\
			$K_s$=400 & 58.33 & -0.79 & 63.86 & 68.38 & \textbf{49.02} & 61.09 \\
			
			\hline
			
		\end{tabular}%
	}
\end{table}

\begin{figure}[htbp]
	\centering
	\includegraphics[width=0.45\textwidth]{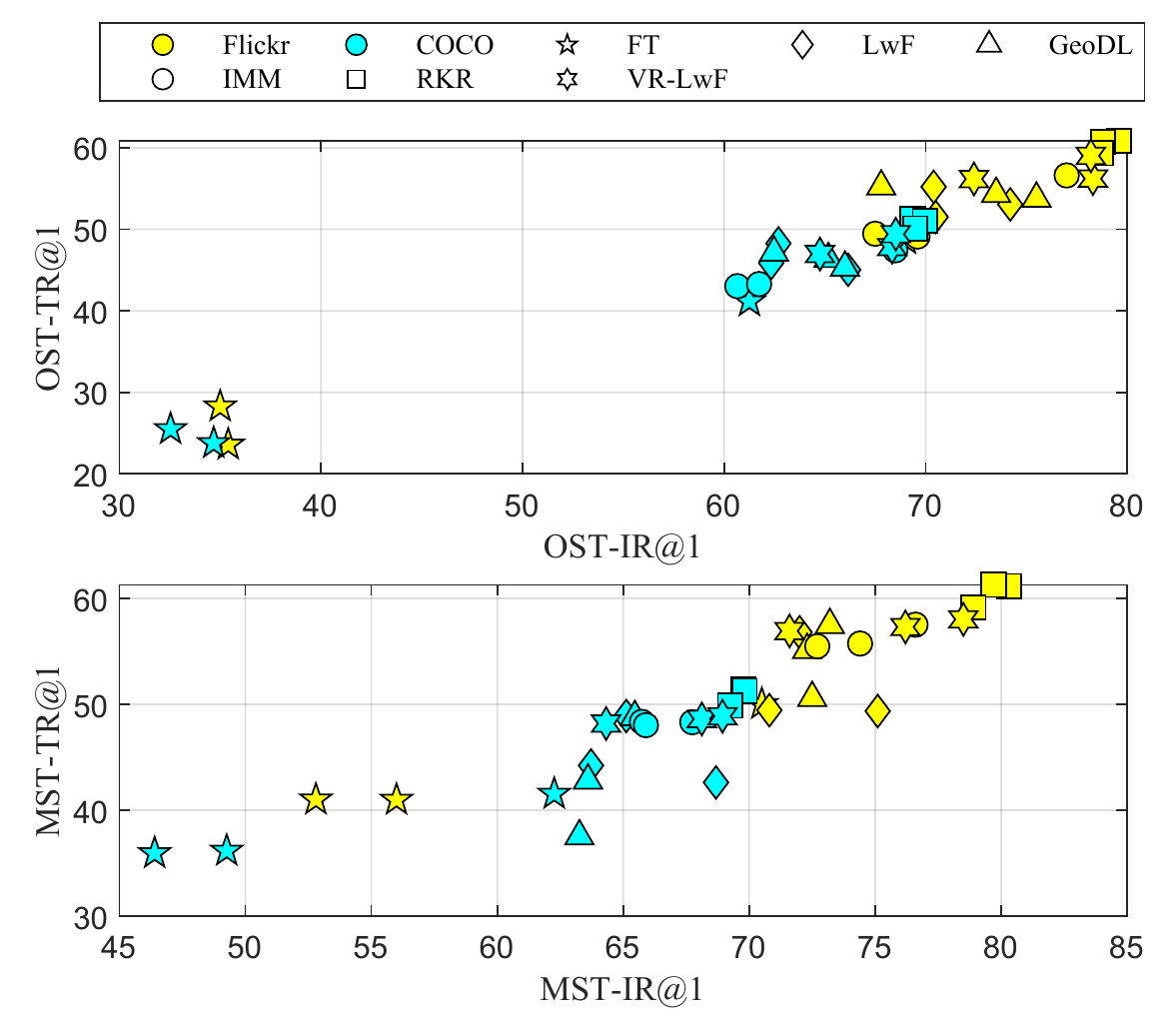}
	\caption{Comparison of Flickr\cite{Flickr30k} and COCO\cite{COCO} retrieval performance, horizontal axis presents TR@1(\%) and vertical axis presents IR@1(\%). The top/bottom chart is for OST/MST results. The figure shows that distributions of Flickr and COCO results are the same, which means the evaluation is transferable.}
	\label{F30K}
\end{figure}

\subsection{Ablations}
In the VR-LwF methods, the pseudo-sentence is generated from a set of randomly sampled words. A crucial question is whether the random replaying is enough? To answer this, we build a caption replaying method, which is randomly sampling image captions from  MSCOCO dataset. This variant ensures that the sampled sentence is a meaningful natural sentence. Once the sentences are sampled, the rest part is identical to the VR-LwF. We report the experimental comparisons in \Cref{M_hyper} and \Cref{MM_hyper}, for OST and MST respectively. Caption-replaying for OST is slightly better on ZS-Acc and retrieval recalls but not very significant. This somehow suggests that grammar does not play a predominant role. As for MST, the caption-replaying is even worse than random sampling. We also examine the effect of sampling length $M$, in \Cref{M_hyper} and \Cref{MM_hyper}, it is seen that $M$ is not influential for the results either. The longer length seems better for MST, but the zero-shot performance of OST drops a little. The drop may be explained by the long random-sampling causes more noise. As for replaying number $K_s$, \Cref{Ks_hyper} and \Cref{MKs_hyper} illustrate that the difference is not distinct. However, the large number leads to a heavy cost of training time.

To evaluate the transferability of the experimental results, we also report the retrieval performance of Flickr30k dataset \cite{Flickr30k}. In \Cref{F30K}, the Flickr results present the same distribution as COCO retrieval, which suggests the generalization of our method and conducted protocol.

\section{Conclusion}
This paper presents the first study on the continual learning problem for large pre-trained CLIP models. We propose an evaluation benchmark with detailed protocols for our study and future research. We explore the extensions of existing continual learning algorithms to address the forgetting issues of CLIP fine-tuning. Our study shows that the CL-CLIP is different from traditional continual learning problem and the model update scheme, \emph{e.g.}, fixing some parts of model parameters, impacts the performance largely. The extension of existing CL methods alleviates the forgetting issue but sacrifices the fine-tuning performance. Moreover, we propose a new method called Learning without Forgetting via Replayed Vocabularies (VR-LwF), which can largely maintain the zero-shot learning capability while making the CLIP model benefit from fine-tuning. To our knowledge, our study complies with general ethical conduct and does not have any ethical issues.

\bibliographystyle{IEEEtran}
\bibliography{egbib}

\begin{thebibliography}{10}
\providecommand{\url}[1]{#1}
\csname url@samestyle\endcsname
\providecommand{\newblock}{\relax}
\providecommand{\bibinfo}[2]{#2}
\providecommand{\BIBentrySTDinterwordspacing}{\spaceskip=0pt\relax}
\providecommand{\BIBentryALTinterwordstretchfactor}{4}
\providecommand{\BIBentryALTinterwordspacing}{\spaceskip=\fontdimen2\font plus
\BIBentryALTinterwordstretchfactor\fontdimen3\font minus
  \fontdimen4\font\relax}
\providecommand{\BIBforeignlanguage}[2]{{%
\expandafter\ifx\csname l@#1\endcsname\relax
\typeout{** WARNING: IEEEtran.bst: No hyphenation pattern has been}%
\typeout{** loaded for the language `#1'. Using the pattern for}%
\typeout{** the default language instead.}%
\else
\language=\csname l@#1\endcsname
\fi
#2}}
\providecommand{\BIBdecl}{\relax}
\BIBdecl

\bibitem{BERT}
J.~Devlin, M.~Chang, K.~Lee, and K.~Toutanova, ``{BERT:} pre-training of deep
  bidirectional transformers for language understanding,'' \emph{CoRR}, vol.
  abs/1810.04805, 2018.

\bibitem{GPT}
A.~Radford, K.~Narasimhan, T.~Salimans, and I.~Sutskever, ``Improving language
  understanding by generative pre-training.''

\bibitem{VilBERT}
J.~Lu, D.~Batra, D.~Parikh, and S.~Lee, ``Vilbert: Pretraining task-agnostic
  visiolinguistic representations for vision-and-language tasks,'' \emph{arXiv
  preprint arXiv:1908.02265}, 2019.

\bibitem{Lxmert}
H.~Tan and M.~Bansal, ``Lxmert: Learning cross-modality encoder representations
  from transformers,'' \emph{arXiv preprint arXiv:1908.07490}, 2019.

\bibitem{CLIP}
A.~Radford, J.~W. Kim, C.~Hallacy, A.~Ramesh, G.~Goh, S.~Agarwal, G.~Sastry,
  A.~Askell, P.~Mishkin, J.~Clark, G.~Krueger, and I.~Sutskever, ``Learning
  transferable visual models from natural language supervision,'' in
  \emph{Proceedings of the 38th International Conference on Machine Learning,
  {ICML} 2021, 18-24 July 2021, Virtual Event}, vol. 139, 2021, pp. 8748--8763.

\bibitem{DetCLIP}
X.~Gu, T.~Lin, W.~Kuo, and Y.~Cui, ``Zero-shot detection via vision and
  language knowledge distillation,'' \emph{CoRR}, vol. abs/2104.13921, 2021.

\bibitem{OpenSetCLIP}
S.~Esmaeilpour, B.~Liu, E.~Robertson, and L.~Shu, ``Zero-shot open set
  detection by extending {CLIP},'' \emph{CoRR}, vol. abs/2109.02748, 2021.

\bibitem{VRCLIP}
X.~Cheng, H.~Lin, X.~Wu, F.~Yang, and D.~Shen, ``Improving video-text retrieval
  by multi-stream corpus alignment and dual softmax loss,'' \emph{CoRR}, vol.
  abs/2109.04290, 2021.

\bibitem{CLIP4Clip}
H.~Luo, L.~Ji, M.~Zhong, Y.~Chen, W.~Lei, N.~Duan, and T.~Li, ``Clip4clip: An
  empirical study of {CLIP} for end to end video clip retrieval,'' \emph{CoRR},
  vol. abs/2104.08860, 2021.

\bibitem{CLIP2Video}
H.~Fang, P.~Xiong, L.~Xu, and Y.~Chen, ``Clip2video: Mastering video-text
  retrieval via image {CLIP},'' \emph{CoRR}, vol. abs/2106.11097, 2021.

\bibitem{LwF}
Z.~Li and D.~Hoiem, ``Learning without forgetting,'' \emph{IEEE transactions on
  pattern analysis and machine intelligence}, vol.~40, no.~12, pp. 2935--2947,
  2017.

\bibitem{GeoDL}
C.~Simon, P.~Koniusz, and M.~Harandi, ``On learning the geodesic path for
  incremental learning,'' in \emph{Proceedings of the IEEE/CVF Conference on
  Computer Vision and Pattern Recognition}, 2021, pp. 1591--1600.

\bibitem{IMM}
S.~Lee, J.~Kim, J.~Jun, J.~Ha, and B.~Zhang, ``Overcoming catastrophic
  forgetting by incremental moment matching,'' in \emph{Advances in Neural
  Information Processing Systems 30: Annual Conference on Neural Information
  Processing Systems 2017, December 4-9, 2017, Long Beach, CA, {USA}}, 2017,
  pp. 4652--4662.

\bibitem{RKR}
P.~Singh, P.~Mazumder, P.~Rai, and V.~P. Namboodiri, ``Rectification-based
  knowledge retention for continual learning,'' in \emph{Proceedings of the
  IEEE/CVF Conference on Computer Vision and Pattern Recognition}, 2021, pp.
  15\,282--15\,291.

\bibitem{videobert}
C.~Sun, A.~Myers, C.~Vondrick, K.~Murphy, and C.~Schmid, ``Videobert: A joint
  model for video and language representation learning,'' in \emph{Proceedings
  of the IEEE/CVF International Conference on Computer Vision}, 2019, pp.
  7464--7473.

\bibitem{UNITER}
Y.~Chen, L.~Li, L.~Yu, A.~E. Kholy, F.~Ahmed, Z.~Gan, Y.~Cheng, and J.~Liu,
  ``{UNITER:} learning universal image-text representations,'' \emph{CoRR},
  vol. abs/1909.11740, 2019.

\bibitem{VisualBERT}
L.~H. Li, M.~Yatskar, D.~Yin, C.~Hsieh, and K.~Chang, ``Visualbert: {A} simple
  and performant baseline for vision and language,'' \emph{CoRR}, vol.
  abs/1908.03557, 2019.

\bibitem{12in1}
J.~Lu, V.~Goswami, M.~Rohrbach, D.~Parikh, and S.~Lee, ``12-in-1: Multi-task
  vision and language representation learning,'' in \emph{2020 {IEEE/CVF}
  Conference on Computer Vision and Pattern Recognition, {CVPR} 2020, Seattle,
  WA, USA, June 13-19, 2020}, 2020, pp. 10\,434--10\,443.

\bibitem{ALIGN}
C.~Jia, Y.~Yang, Y.~Xia, Y.~Chen, Z.~Parekh, H.~Pham, Q.~V. Le, Y.~Sung, Z.~Li,
  and T.~Duerig, ``Scaling up visual and vision-language representation
  learning with noisy text supervision,'' in \emph{Proceedings of the 38th
  International Conference on Machine Learning, {ICML} 2021, 18-24 July 2021,
  Virtual Event}, vol. 139, 2021, pp. 4904--4916.

\bibitem{WenLan}
Y.~Huo, M.~Zhang, G.~Liu, H.~Lu, Y.~Gao, G.~Yang, J.~Wen, H.~Zhang, B.~Xu,
  W.~Zheng, Z.~Xi, Y.~Yang, A.~Hu, J.~Zhao, R.~Li, Y.~Zhao, L.~Zhang, Y.~Song,
  X.~Hong, W.~Cui, D.~Y. Hou, Y.~Li, J.~Li, P.~Liu, Z.~Gong, C.~Jin, Y.~Sun,
  S.~Chen, Z.~Lu, Z.~Dou, Q.~Jin, Y.~Lan, W.~X. Zhao, R.~Song, and J.~Wen,
  ``Wenlan: Bridging vision and language by large-scale multi-modal
  pre-training,'' \emph{CoRR}, vol. abs/2103.06561, 2021.

\bibitem{forgetting}
M.~McCloskey and N.~J. Cohen, ``Catastrophic interference in connectionist
  networks: The sequential learning problem,'' in \emph{Psychology of learning
  and motivation}.\hskip 1em plus 0.5em minus 0.4em\relax Elsevier, 1989,
  vol.~24, pp. 109--165.

\bibitem{EWC}
J.~Kirkpatrick, R.~Pascanu, N.~Rabinowitz, J.~Veness, G.~Desjardins, A.~A.
  Rusu, K.~Milan, J.~Quan, T.~Ramalho, A.~Grabska-Barwinska \emph{et~al.},
  ``Overcoming catastrophic forgetting in neural networks,'' \emph{Proceedings
  of the national academy of sciences}, vol. 114, no.~13, pp. 3521--3526, 2017.

\bibitem{LC}
D.-W. Zhou, Y.~Yang, and D.-C. Zhan, ``Learning to classify with incremental
  new class,'' \emph{IEEE Transactions on Neural Networks and Learning
  Systems}, 2021.

\bibitem{IRL}
Z.~Wang, H.-X. Li, and C.~Chen, ``Incremental reinforcement learning in
  continuous spaces via policy relaxation and importance weighting,''
  \emph{IEEE transactions on neural networks and learning systems}, vol.~31,
  no.~6, pp. 1870--1883, 2019.

\bibitem{OIC}
J.-Y. Park and J.-H. Kim, ``Online incremental classification resonance network
  and its application to human--robot interaction,'' \emph{IEEE transactions on
  neural networks and learning systems}, vol.~31, no.~5, pp. 1426--1436, 2019.

\bibitem{OEWC}
J.~Schwarz, W.~Czarnecki, J.~Luketina, A.~Grabska-Barwinska, Y.~W. Teh,
  R.~Pascanu, and R.~Hadsell, ``Progress \& compress: A scalable framework for
  continual learning,'' in \emph{International Conference on Machine Learning},
  2018, pp. 4528--4537.

\bibitem{MAS}
R.~Aljundi, F.~Babiloni, M.~Elhoseiny, M.~Rohrbach, and T.~Tuytelaars, ``Memory
  aware synapses: Learning what (not) to forget,'' in \emph{Proceedings of the
  European Conference on Computer Vision (ECCV)}, 2018, pp. 139--154.

\bibitem{SI}
F.~Zenke, B.~Poole, and S.~Ganguli, ``Continual learning through synaptic
  intelligence,'' in \emph{International Conference on Machine Learning}, 2017,
  pp. 3987--3995.

\bibitem{NPC}
I.~Paik, S.~Oh, T.~Kwak, and I.~Kim, ``Overcoming catastrophic forgetting by
  neuron-level plasticity control,'' in \emph{Proceedings of the AAAI
  Conference on Artificial Intelligence}, vol.~34, no.~04, 2020, pp.
  5339--5346.

\bibitem{Laplacian}
D.~J. MacKay, ``A practical bayesian framework for backpropagation networks,''
  \emph{Neural computation}, vol.~4, no.~3, pp. 448--472, 1992.

\bibitem{KD}
G.~E. Hinton, O.~Vinyals, and J.~Dean, ``Distilling the knowledge in a neural
  network,'' \emph{CoRR}, vol. abs/1503.02531, 2015.

\bibitem{LwM}
P.~Dhar, R.~V. Singh, K.-C. Peng, Z.~Wu, and R.~Chellappa, ``Learning without
  memorizing,'' in \emph{Proceedings of the IEEE/CVF Conference on Computer
  Vision and Pattern Recognition}, 2019, pp. 5138--5146.

\bibitem{Podnet}
A.~Douillard, M.~Cord, C.~Ollion, T.~Robert, and E.~Valle, ``Podnet: Pooled
  outputs distillation for small-tasks incremental learning,'' in
  \emph{Computer Vision--ECCV 2020: 16th European Conference, Glasgow, UK,
  August 23--28, 2020, Proceedings, Part XX 16}, 2020, pp. 86--102.

\bibitem{MCIL}
Y.~Liu, Y.~Su, A.-A. Liu, B.~Schiele, and Q.~Sun, ``Mnemonics training:
  Multi-class incremental learning without forgetting,'' in \emph{Proceedings
  of the IEEE/CVF conference on Computer Vision and Pattern Recognition}, 2020,
  pp. 12\,245--12\,254.

\bibitem{Grad-cam}
R.~R. Selvaraju, M.~Cogswell, A.~Das, R.~Vedantam, D.~Parikh, and D.~Batra,
  ``Grad-cam: Visual explanations from deep networks via gradient-based
  localization,'' in \emph{Proceedings of the IEEE international conference on
  computer vision}, 2017, pp. 618--626.

\bibitem{SER}
D.~Isele and A.~Cosgun, ``Selective experience replay for lifelong learning,''
  in \emph{Proceedings of the AAAI Conference on Artificial Intelligence},
  vol.~32, no.~1, 2018.

\bibitem{iCARL}
S.-A. Rebuffi, A.~Kolesnikov, G.~Sperl, and C.~H. Lampert, ``icarl: Incremental
  classifier and representation learning,'' in \emph{Proceedings of the IEEE
  conference on Computer Vision and Pattern Recognition}, 2017, pp. 2001--2010.

\bibitem{Gdumb}
A.~Prabhu, P.~H. Torr, and P.~K. Dokania, ``Gdumb: A simple approach that
  questions our progress in continual learning,'' in \emph{European conference
  on computer vision}, 2020, pp. 524--540.

\bibitem{Generative}
H.~Shin, J.~K. Lee, J.~Kim, and J.~Kim, ``Continual learning with deep
  generative replay,'' in \emph{Advances in Neural Information Processing
  Systems 30: Annual Conference on Neural Information Processing Systems 2017,
  December 4-9, 2017, Long Beach, CA, {USA}}, 2017, pp. 2990--2999.

\bibitem{VAECL}
F.~Lavda, J.~Ramapuram, M.~Gregorova, and A.~Kalousis, ``Continual
  classification learning using generative models,'' \emph{CoRR}, vol.
  abs/1810.10612, 2018.

\bibitem{TMN}
L.~Wang, B.~Lei, Q.~Li, H.~Su, J.~Zhu, and Y.~Zhong, ``Triple-memory networks:
  A brain-inspired method for continual learning,'' \emph{IEEE Transactions on
  Neural Networks and Learning Systems}, vol.~33, no.~5, pp. 1925--1934, 2021.

\bibitem{GAN}
I.~Goodfellow, J.~Pouget-Abadie, M.~Mirza, B.~Xu, D.~Warde-Farley, S.~Ozair,
  A.~Courville, and Y.~Bengio, ``Generative adversarial nets,'' \emph{Advances
  in neural information processing systems}, vol.~27, 2014.

\bibitem{VAE}
D.~P. Kingma and M.~Welling, ``Auto-encoding variational bayes,'' \emph{arXiv
  preprint arXiv:1312.6114}, 2013.

\bibitem{DER}
S.~Yan, J.~Xie, and X.~He, ``Der: Dynamically expandable representation for
  class incremental learning,'' in \emph{Proceedings of the IEEE/CVF Conference
  on Computer Vision and Pattern Recognition}, 2021, pp. 3014--3023.

\bibitem{DEN}
J.~Yoon, E.~Yang, J.~Lee, and S.~J. Hwang, ``Lifelong learning with dynamically
  expandable networks,'' in \emph{6th International Conference on Learning
  Representations, {ICLR} 2018, Vancouver, BC, Canada, April 30 - May 3, 2018,
  Conference Track Proceedings}, 2018.

\bibitem{DM}
G.-M. Park, S.-M. Yoo, and J.-H. Kim, ``Convolutional neural network with
  developmental memory for continual learning,'' \emph{IEEE Transactions on
  Neural Networks and Learning Systems}, vol.~32, no.~6, pp. 2691--2705, 2020.

\bibitem{HAT}
J.~Serra, D.~Suris, M.~Miron, and A.~Karatzoglou, ``Overcoming catastrophic
  forgetting with hard attention to the task,'' in \emph{International
  Conference on Machine Learning}, 2018, pp. 4548--4557.

\bibitem{WiSE-FT}
M.~Wortsman, G.~Ilharco, M.~Li, J.~W. Kim, H.~Hajishirzi, A.~Farhadi,
  H.~Namkoong, and L.~Schmidt, ``Robust fine-tuning of zero-shot models,''
  \emph{CoRR}, vol. abs/2109.01903, 2021.

\bibitem{COCO}
T.-Y. Lin, M.~Maire, S.~Belongie, J.~Hays, P.~Perona, D.~Ramanan,
  P.~Doll{\'a}r, and C.~L. Zitnick, ``Microsoft coco: Common objects in
  context,'' in \emph{European conference on computer vision}, 2014, pp.
  740--755.

\bibitem{karpathy}
A.~Karpathy and L.~Fei-Fei, ``Deep visual-semantic alignments for generating
  image descriptions,'' in \emph{Proceedings of the IEEE conference on computer
  vision and pattern recognition}, 2015, pp. 3128--3137.

\bibitem{teired-ImageNet}
M.~Ren, E.~Triantafillou, S.~Ravi, J.~Snell, K.~Swersky, J.~B. Tenenbaum,
  H.~Larochelle, and R.~S. Zemel, ``Meta-learning for semi-supervised few-shot
  classification,'' in \emph{6th International Conference on Learning
  Representations, {ICLR} 2018, Vancouver, BC, Canada, April 30 - May 3, 2018,
  Conference Track Proceedings}, 2018.

\bibitem{Imagenet}
O.~Russakovsky, J.~Deng, H.~Su, J.~Krause, S.~Satheesh, S.~Ma, Z.~Huang,
  A.~Karpathy, A.~Khosla, M.~Bernstein \emph{et~al.}, ``Imagenet large scale
  visual recognition challenge,'' \emph{International journal of computer
  vision}, vol. 115, no.~3, pp. 211--252, 2015.

\bibitem{Transformer}
A.~Vaswani, N.~Shazeer, N.~Parmar, J.~Uszkoreit, L.~Jones, A.~N. Gomez,
  L.~Kaiser, and I.~Polosukhin, ``Attention is all you need,'' in
  \emph{Advances in Neural Information Processing Systems 30: Annual Conference
  on Neural Information Processing Systems 2017, December 4-9, 2017, Long
  Beach, CA, {USA}}, 2017, pp. 5998--6008.

\bibitem{VSE}
F.~Faghri, D.~J. Fleet, J.~R. Kiros, and S.~Fidler, ``Vse++: Improving
  visual-semantic embeddings with hard negatives,'' \emph{arXiv preprint
  arXiv:1707.05612}, 2017.

\bibitem{Adam}
D.~P. Kingma and J.~Ba, ``Adam: {A} method for stochastic optimization,'' in
  \emph{3rd International Conference on Learning Representations, {ICLR} 2015,
  San Diego, CA, USA, May 7-9, 2015, Conference Track Proceedings}, 2015.

\bibitem{GEM}
D.~Lopez-Paz and M.~Ranzato, ``Gradient episodic memory for continual
  learning,'' \emph{Advances in neural information processing systems},
  vol.~30, pp. 6467--6476, 2017.

\bibitem{Flickr30k}
B.~A. Plummer, L.~Wang, C.~M. Cervantes, J.~C. Caicedo, J.~Hockenmaier, and
  S.~Lazebnik, ``Flickr30k entities: Collecting region-to-phrase
  correspondences for richer image-to-sentence models,'' \emph{International
  Journal of Computer Vision}, vol. 123, no.~1, pp. 74--93, 2017.

\end{thebibliography}

\end{document}